\documentclass[11pt]{article}

\usepackage[final]{acl}

\usepackage{times}
\usepackage{latexsym}

\usepackage{multirow}
\usepackage{booktabs}

\usepackage{xcolor}
\usepackage{tcolorbox}

\usepackage{array}
\usepackage{tabularx}
\usepackage{ragged2e}
\usepackage{pifont}
\usepackage{enumitem}

\tcbuselibrary{breakable}
\definecolor{boxgray}{gray}{0.95}
\newtcolorbox{promptbox}[1]{
  breakable,
  colback=boxgray,
  colframe=black,
  colbacktitle=black,
  coltitle=white,
  fonttitle=\bfseries\large,
  fontupper=\rmfamily,
  title={#1},
  arc=1.5mm,
  boxrule=1pt,
  left=3mm, right=3mm, top=3mm, bottom=3mm,
  toptitle=2mm, bottomtitle=2mm
}
\newcolumntype{L}{>{\RaggedRight\arraybackslash}X}
\newcolumntype{C}{>{\Centering\arraybackslash}X}

\usepackage[T1]{fontenc}

\usepackage[utf8]{inputenc}

\usepackage{microtype}

\usepackage{inconsolata}

\usepackage{graphicx}
\usepackage{amsmath}
\usepackage{amsfonts}
\usepackage{algorithm}
\usepackage{algpseudocode}
\title{What Deserves Memory: Adaptive Memory Distillation for LLM Agents}

\author{
  \textbf{Wenquan Ma\textsuperscript{1, 4}\thanks{Equal contribution.}},
  \textbf{Jiayan Nan\textsuperscript{2}\footnotemark[1]\thanks{Corresponding author.}},
  \textbf{Wenlong Wu\textsuperscript{3}},
  \textbf{Yize Chen\textsuperscript{2}}
\\
\\
  \textsuperscript{1}Fudan University\quad
  \textsuperscript{2}Shanda Group\quad
  \textsuperscript{3}Beihang University\quad
\\
  \textsuperscript{4}Shanghai University of Finance and Economics
\\
  \small{
    \texttt{wenquan.ma@outlook.com},
    \texttt{nanjiayan@shanda.com},
    \texttt{wlw@buaa.edu.cn}
    \texttt{chenyize@shanda.com}
  }
}

\begin{document}
\maketitle
\begin{abstract}
Memory systems for LLM agents struggle to determine what information deserves retention. 
Existing approaches rely on predefined heuristics such as importance scores, emotional tags, or factual templates, encoding designer intuition rather than learning from the data itself. 
Inspired by cognitive ideas, we propose \textbf{\textsc{Nemori}}, an adaptive memory distillation framework that casts the assessment of the experience's future utility as a matter of predictability. 
Specifically, \textsc{Nemori} comprises two cascading modules: Episodic Memory Integration transforms raw interactions into coherent narratives, and Semantic Knowledge Distillation extracts insights via prediction error. 
Centering on distillation, the framework remains agnostic to downstream management. 
Extensive experiments confirm that \textsc{Nemori} achieves strong performance, efficiency, and storage reduction. 
Our work suggests that observing the intrinsic properties of interaction sequences offers a viable, data-driven alternative to heuristic-based memory design. 
Code at: \url{https://github.com/nemori-ai/nemori}.
\end{abstract}

\section{Introduction}
\label{sec:intro}
The difficulty of maintaining long-term behavioral consistency in Large Language Model (LLM)-based agents stems from a fundamental conflict: the reliance of stateless LLMs on linearly expanding interaction trajectories versus the constraints of finite context window and the \emph{Lost in the Middle} phenomenon \citep{liu-etal-2024-lost}.
Nevertheless, real-world applications, exemplified by personal assistants, autonomous agents and personalized recommendation systems, increasingly demand persistent interaction. 
To address these challenges, memory systems that facilitate real-time context regulation have emerged as a viable and prevailing approach. 

\begin{table*}[t]
    \centering
    \small
    \renewcommand{\arraystretch}{1.2}
    \begin{tabularx}{\textwidth}{@{} l l C C C @{}}
    \toprule
    \textbf{Category} & \textbf{Method} & \textbf{Distillation} & \textbf{Management} & \textbf{Retrieval} \\ \midrule
    
    \textit{Retrieval-time} 
    & \citet{DBLP:conf/nips/LewisPPPKGKLYR020} & --- & --- & Similarity search \\ \hline
    
    \multirow{7}{*}{\textit{Management-time}} 
    & \citet{DBLP:journals/corr/abs-2310-08560} & --- & Tiered storage & Function calls \\
    & \citet{DBLP:conf/aaai/ZhongGGYW24} & --- & Summary + forgetting & --- \\
    & \citet{kang-etal-2025-memory} & --- & Heat scoring & Two-step search \\
    & \citet{li2025cam} & --- & Hierarchical summary & Multi-step search \\
    & \citet{DBLP:conf/ijcai/AnokhinSSEK0B25} & --- & Graph update & Graph spreading \\
    & \citet{xu2025amem} & --- & Adaptive note linking & --- \\
    & \citet{DBLP:journals/corr/abs-2501-13956} & --- & Validity management & Reranking \\ \hline
    
    \multirow{5}{*}{\textit{Distillation-time}} 
    & \citet{DBLP:conf/uist/ParkOCMLB23} & Importance scoring & Reflection trees & Weighted scoring \\
    & \citet{DBLP:conf/icbk/HuangLS0B24} & Emotion tagging & --- & Emotion matching \\
    & \citet{DBLP:journals/corr/abs-2504-19413} & Facts extraction & --- & --- \\
    & \citet{li-etal-2025-hello} & Summary + persona & --- & Noun overlap \\
    & \citet{DBLP:conf/iclr/PanWJLCL0LZQ025} & Topic & Token compression & --- \\ \midrule
    
    & \textbf{This paper} & \textbf{Prediction error} & \textbf{Agnostic} & \textbf{---} \\ \bottomrule
    
    \end{tabularx}
    \caption{Agent memory systems categorized by the stage at which memory utility is assessed. Retrieval-time methods defer assessment entirely; management-time methods filter post-hoc via access patterns; distillation-time methods assess at entry ingestion. Common practices omitted: raw retention in Distillation, conflict detection in Management, similarity search (non-graph) or graph traversal (graph-based) in Retrieval. Our approach assesses at distillation via prediction error rather than predefined heuristics.}
    \label{tab:memory_taxonomy}
\end{table*}

Memory systems identify useful experiences to facilitate future response generation through two stages: \emph{distillation}, which determines the entry form of experiences, and \emph{management}, which ensures their ongoing maintenance. 
To this end, one category of approaches focuses on management by treating entries as opaque containers, where utility is inferred through observable structural metadata, such as access frequency \citep{kang-etal-2025-memory}, temporal decay \citep{DBLP:conf/aaai/ZhongGGYW24}, or explicit relationships \citep{xu2025amem}, foregoing the inspection of the nuanced content itself.
In contrast, another category of approaches intervenes during the initial distillation stage, selectively shaping the entry form. 
This \emph{pre-positioning}, while granting greater flexibility, must contend with future utility uncertainty.
Existing distillation methods typically address this by encoding designer intuition, such as importance scores \citep{DBLP:conf/uist/ParkOCMLB23}, emotional tags \citep{DBLP:conf/icbk/HuangLS0B24}, or factual templates \citep{DBLP:journals/corr/abs-2504-19413}. 
However, such heuristics risk introducing subjective bias, which is fatal during distillation as it can lead to irreversible information distortion, or causing systemic bloat, where the system tends to over-store to avoid such distortion, thereby amplifying retrieval noise. 
This limitation necessitates an approach that assesses the potential utility grounded in the interaction experience itself.

Inspired by Predictive Coding Theory \citep{rao1999predictive, friston_free-energy_2010, clark2013whatever}, we propose \textbf{\textsc{Nemori}}, a training-free framework that casts the assessment of experience utility as adaptive memory distillation over incoming observations that the agent fails to predict given existing knowledge, enabling a data-driven space.
As illustrated in Figure~\ref{fig:architecture}, this framework, guided by three parsimonious priors over memory structure, representation and distillation, comprises two cascading modules, echoing Complementary Learning Systems \citep{mcclelland1995there}. 
Specifically, the \emph{Episodic Memory Integration} module first transforms raw interaction sequences into coherent episodic narratives. 
The \emph{Semantic Knowledge Distillation} module then extracts novel experience that existing knowledge cannot anticipate. 
Centering on the distillation stage, \textsc{Nemori} remains agnostic to the underlying management, while a native management system is provided.
Our contributions:

1) \textbf{Perspective.} We formalize the distinction between \emph{distillation} and \emph{management} in memory construction, and derive priors from general data properties and cognitive ideas to guide distillation design.

2) \textbf{Framework.} We implement \textsc{Nemori}, a management-agnostic adaptive memory distillation framework, and equip it with a native management system.

3) \textbf{Evaluation.} We conduct extensive experiments demonstrating \textsc{Nemori}'s strong performance, with pronounced advantages in longer context. When integrated with third-party management systems, \textsc{Nemori} enhances A-MEM and MemoryOS with 45--64\% storage reduction while maintaining performance. 

\section{Related Works}
\subsection{Memory for LLM Agents}
Agent memory systems decompose into three stages: distillation (what to retain), management (how to organize), and retrieval (how to surface content). 
Unlike pure RAG \citep{DBLP:conf/nips/LewisPPPKGKLYR020} that defers judgment to query time, memory systems \emph{pre-position}: enriching data with metadata at distillation, then utilizing it for management and retrieval. 
Table~\ref{tab:memory_taxonomy} categorizes works by when enrichment occurs. 
\emph{Management-time} methods enrich post-hoc via decay weights, tiered storage, access frequency, or relationship linking \citep{DBLP:conf/aaai/ZhongGGYW24, DBLP:journals/corr/abs-2310-08560, kang-etal-2025-memory, xu2025amem}. 
\emph{Distillation-time} methods enrich at ingestion through importance scoring, emotional tagging, or fact extraction \citep{DBLP:conf/uist/ParkOCMLB23, DBLP:conf/icbk/HuangLS0B24, DBLP:journals/corr/abs-2504-19413}. 

\begin{figure*}[t]
\centering
\includegraphics[width=\textwidth]{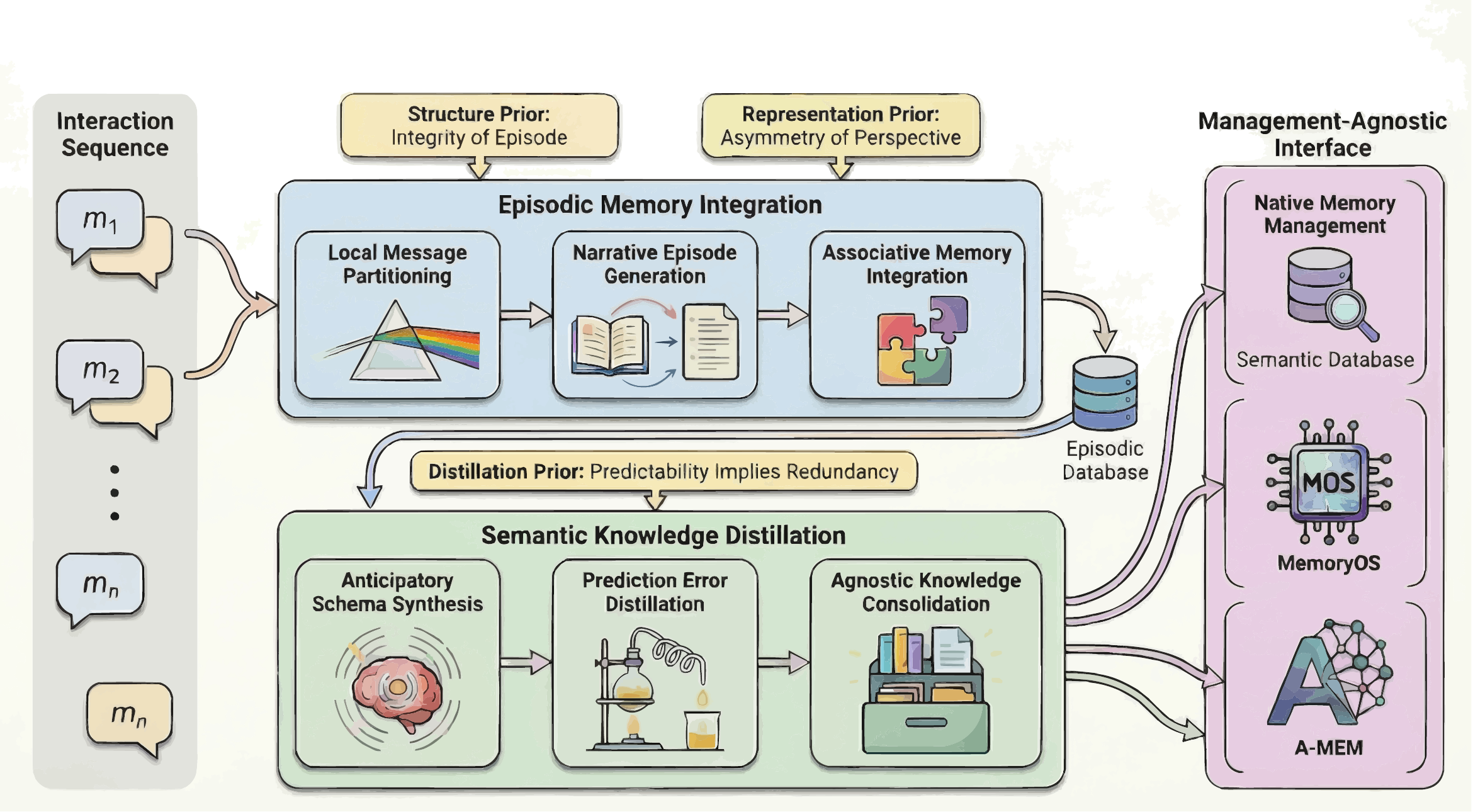}
\caption{Overview of the \textsc{Nemori} framework. The system comprises two cascading modules guided by three priors: Episodic Memory Integration (top) transforms raw interactions into coherent narrative episodes, and Semantic Knowledge Distillation (bottom) extracts insights via prediction error. The framework can serve as a distillation layer complement native or third-party management systems (right).}
\label{fig:architecture}
\end{figure*}

\subsection{Cognitive Principles of Memory}
Predictive Coding Theory \citep{rao1999predictive}, originally from visual neuroscience, posits that higher cortical areas send predictions downward while lower areas propagate primarily the residual \emph{prediction error} upward.
\citet{friston_free-energy_2010} generalized this into the Free Energy Principle, a unifying framework across perception, action, and learning. 
\citet{clark2013whatever} further extended it, arguing that brains are fundamentally prediction machines. 
\textsc{Nemori} adapts this insight to agent memory design: prediction error signals information worth retaining; what is predictable is therefore redundant.

\section{Methodology}
\label{sec:method}
\textsc{Nemori} is an adaptive memory distillation framework inspired by cognitive ideas \citep{mcclelland1995there, rao1999predictive, friston_free-energy_2010, clark2013whatever}. 
This management-agnostic framework can serve as a distillation layer that complements either native or third-party memory systems like A-MEM or MemoryOS. 
A production-grade implementation is provided at \url{https://github.com/nemori-ai/nemori}. 

\subsection{Overview \& Motivations}
As illustrated in Figure~\ref{fig:architecture}, \textsc{Nemori} comprises two cascading modules guided by three priors as inductive biases. These priors capture the parsimonious features of continuous interaction sequences, establishing a plastic, data-driven environment in which intrinsic dynamics drive the partitioning, representation, and distillation of experience into memory. 

\paragraph{The Structure Prior: Integrity of Episode.}
Interaction sequences exhibit natural grouping. 
Interactions within each episodic group are mutually contextualizing: individual messages derive their meaning, partly, from surrounding ones, and finer-grained or arbitrary fragmentation would sever the context that renders them interpretable. 

This prior requires the framework to define episodes respecting latent integrity among interactions, rather than imposing heuristic chunking. 

\paragraph{The Representation Prior: Asymmetry of Perspective.}
Memory serves recall. 
Recalling is essentially a form of reasoning, an allocentric reconstruction of events, whereas raw episodes are egocentric and inherently noisy. 

This prior requires the framework to transform raw episodes into narrative representations that highlight logical structures while preserving salient details, bridging the gap between chaotic perception and rational retrieval. 
 
\paragraph{The Distillation Prior: Predictability Implies Redundancy.}
Information within interaction sequences is highly redundant. 
From the perspective of predictive coding, the unexpected information is a natural candidate for memory consolidation. 

This key prior requires the framework to distill memory by inspecting the semantic differential between the actual interactions and their anticipatory schema derived from existing knowledge. 

In the following sections, we detail the framework implementation guided by these priors. 

\subsection{Episodic Memory Integration}
\label{subsec: Episodic}
Guided by the structure prior and the representation prior, this module integrates raw interactions into episodic memories and prepares them for subsequent distillation. 
It is further divided into three submodules: Local Message Partitioning, Narrative Episode Generation and Associative Memory Integration. 

\subsubsection{Local Message Partitioning}
\label{sec:partitioning}
Guided by the structure prior, this submodule resolves the continuous interaction sequence within an observation window into a discrete partition. 
We model the interactions between an agent and its environment as a sequence of message exchanges, maintaining a dedicated message buffer $\mathcal{B}$. 
At any time $t$, the buffer state is represented as a queue of messages $\mathcal{B}_t = \{m_1, m_2, \dots, m_z\}$, where each message $m_i = (r_i, c_i, \tau_i)$ specifies the sender, content and timestamp, respectively. 
New interactions are appended to the rear of $\mathcal{B}_t$ as they occur. 

The partitioning process is triggered once the buffer size $|\mathcal{B}_t|$ reaches a predefined observation window length $w \in \mathbb{Z}^{+}$. 
At this juncture, the submodule performs a partitioning operation:
\begin{equation*}
    \mathbf{O} \leftarrow f_{\text{LLM}}(\mathcal{P}_{\text{par}} \parallel \mathcal{B}_t),
\end{equation*}
where $\mathcal{P}_{\text{par}}$ is a prompt that instructs the LLM to discern the latent integrity and local nuances within the window and partition the messages accordingly. 
The output $\mathbf{O} = \{O_1, O_2, \dots, O_n\}$ (where $n \le w$) constitutes a partition of the index set $\{1, 2, \dots, w\}$. 
Specifically, the $O_j$ are pairwise disjoint and their union covers the index set. 
The submodule then maps these indices back to the message buffer to form a collection of raw episodes $\mathbf{P} = \{P_1, \dots, P_n\}$, where each $P_j$ is the subsequence of $\mathcal{B}_t$ indexed by $O_j$. 

Finally, $\mathbf{P}$ is transferred to the Narrative Episode Generation submodule, and the buffer $\mathcal{B}$ is reset to empty to await subsequent incoming messages. 

\subsubsection{Narrative Episode Generation}
\label{sec:narrative}
Guided by the representation prior, this submodule transforms the received raw episodes into narrative representations. 
For each raw episode $P_j \in \mathbf{P}$, the submodule generates a narrative episode $N_j$ and a corresponding episodic cue $c_j$ tailored for semantic distillation:
\begin{equation*}
    (N_j, c_j) \leftarrow f_{\text{LLM}}(\mathcal{P}_{\text{nar}} \parallel P_j),
\end{equation*}
where $\mathcal{P}_{\text{nar}}$ instructs the LLM to highlight the logical structure and constituent elements within the interaction. 

Subsequently, the submodule computes an embedding index $\mathbf{v}_j$ to enable associative retrieval:
\begin{equation*}
    \mathbf{v}_j \leftarrow f_{\text{emb}}(c_j \parallel N_j),
\end{equation*}
where $f_{\text{emb}}$ denotes the embedding model. 

Each episodic memory is represented as $M_j = (c_j, N_j, P_j, \mathbf{v}_j)$. Finally, the collection $\{M_j\}$ is transferred to the Associative Memory Integration submodule.

\paragraph{Discussion.}
Two aspects of this design merit attention. First, it enables dual-mode retrieval: returning $N$ directly for efficiency, or returning raw $P$ for precision-critical domains. Second, from this point onward, the episode becomes the primary processing unit throughout the pipeline, avoiding the \emph{message-wise} processing that many baselines fall into and that incurs substantial cost overhead (see Section~\ref{subsec:efficiency}).

\subsubsection{Associative Memory Integration}
\label{sec:integration}
This submodule dynamically integrates episodes that may have been sundered by the constraints of the observation window length. 
For each new episodic memory $M_j$, the submodule performs an integration check against the existing episodic database $\mathcal{D}_e$. 
It first retrieves the top $K_e$ candidates based on cosine similarity:
\begin{equation*}
    \mathbf{C} = \{U_1, U_2, \dots, U_{K_e}\} \leftarrow \operatorname{Search}(\mathcal{D}_e, \mathbf{v}_j, K_e),
\end{equation*}
where each candidate $U_k = (c_k, N_k, P_k, \mathbf{v}_k)$.

The submodule then selects the optimal integration target:
\begin{equation*}
    idx \leftarrow f_{\text{LLM}}(\mathcal{P}_{\text{sel}} \parallel (c_j, N_j) \parallel \{(c_k, N_k)\}_{k=1}^{K_e}),
\end{equation*}
where $\mathcal{P}_{\text{sel}}$ instructs the LLM to identify the target candidate that shares episodic continuity with the new memory. 
The output $idx \in \{1, \dots, K_e\} \cup \{-1\}$ determines the subsequent operation:

\textbf{Case 1 ($idx=k$):} The LLM integrates two memories: $(c_{\nu}, N_{\nu}) \leftarrow f_{\text{LLM}}(\mathcal{P}_{\text{int}} \parallel c_{k} \parallel N_{k} \parallel c_j \parallel N_j)$, superseding $U_{k}$ with $M_{\nu} = (c_{\nu}, N_{\nu}, P_{k} \parallel P_j, f_{\text{emb}}(c_{\nu} \parallel N_{\nu}))$.

\textbf{Case 2 ($idx=-1$):} No continuity found; $M_j$ is inserted as a distinct entry. 

Finally, the resulting episodic memory ($M_{\nu}$ or $M_j$) is transferred to the Semantic Knowledge Distillation module. 

\subsection{Semantic Knowledge Distillation}
\label{subsec: Semantic}
Guided by the distillation prior, this module implements a management-agnostic process to distill semantic knowledge from episodic experiences by defining generic interfaces for context evocation and knowledge consolidation. 
This process comprises three submodules: Anticipatory Schema Synthesis, Prediction Error Distillation, Agnostic Knowledge Consolidation. 

\subsubsection{Anticipatory Schema Synthesis}
\label{subsec:predict}
This submodule synthesizes an anticipatory schema for each incoming episode by orchestrating existing knowledge. 
We treat the underlying management system $\mathcal{M}$ as an abstract context provider. 
Let $M_{in}$ denote the input episodic memory, newly formed or integrated alike. 
The submodule first invokes a generic interface $\operatorname{Evoke}(\cdot)$ to evoke the context $\mathcal{S}_{in}$ pertaining to $M_{in}$ from $\mathcal{M}$:
\begin{equation*} 
    \mathcal{S}_{in} \leftarrow \operatorname{Evoke}(M_{in}, \mathcal{M}).
\end{equation*}

In our native implementation, this is realized as threshold-filtered similarity search: $\mathcal{S}_{in} \leftarrow \operatorname{Top-}K_s \left( { S_r \in \mathcal{D}_s \mid \operatorname{sim}(\mathbf{v}_{in}, \mathbf{u}_r) > \tau } \right)$.
Alternative variants are detailed in Appendix \ref{app: Instantiations}. 

The anticipatory schema $\hat{P}_{in}$ is then synthesized given only the episodic cue $c_{in}$ (a brief summary of the incoming episode) and the evoked context $\mathcal{S}_{in}$ (what the system already knows):
\begin{equation*}
    \hat{P}_{in} \leftarrow f_{\text{LLM}}(\mathcal{P}_{\text{ant}} \parallel c_{in} \parallel \mathcal{S}_{in}),
\end{equation*}
where $\mathcal{P}_{\text{ant}}$ instructs the LLM to \emph{predict what actually happened} in the incoming episode based on the provided context. 
The resulting $\hat{P}_{in}$ represents the system's guess of the episode content from existing knowledge alone. 

\subsubsection{Prediction Error Distillation}
\label{subsec:distill}
This submodule distills semantic insights from the discrepancy between the raw episode and the anticipatory schema. The process is defined as:
\begin{equation*}
    \mathcal{K}_{in} = \{k_1, \dots, k_d\} \leftarrow f_{\text{LLM}}(\mathcal{P}_{\text{dis}} \parallel P_{in} \parallel \hat{P}_{in}),
\end{equation*}
where $\mathcal{P}_{\text{dis}}$ instructs the LLM to identify and extract, as semantic insights $\mathcal{K}_{in}$, information in the raw episode $P_{in}$ that deviates from or extends the anticipatory schema $\hat{P}_{in}$.

\subsubsection{Agnostic Knowledge Consolidation}
This submodule consolidates the distilled $\mathcal{K}_{in}$ into the underlying management system $\mathcal{M}$, defined by a generic interface $\operatorname{Consolidate}(\mathcal{K}_{in}, \mathcal{M})$, with diverse implementations detailed in Appendix \ref{app: Instantiations}. 

In our native implementation, for each distilled insight $k_q \in \mathcal{K}_{in}$, the submodule first retrieves its associative knowledge $\tilde{\mathcal{S}}_q$ from the semantic database $\mathcal{D}_s$ based on the embedding $\mathbf{u}_q \leftarrow f_{\text{emb}}(k_q)$: 

\begin{equation*}
    \tilde{\mathcal{S}}_q =\{(k_h, \mathbf{u}_h)\}_{h=1}^{K_m} \leftarrow \operatorname{Search}(\mathcal{D}_s, \mathbf{u}_q, K_m).
\end{equation*}

The submodule then resolves the relationship between the insight and existing knowledge, generating a consolidation directive: 
\begin{equation*} 
(\delta, idxs, k_{\mu}) \leftarrow f_{\text{LLM}}(\mathcal{P}_{\text{con}} \parallel k_q \parallel \{k_h\}_{h=1}^{K_m}),
\end{equation*}
where $\mathcal{P}_{\text{con}}$ instructs the LLM to determine consolidation operations, $\delta \in \{\text{\texttt{new}}, \text{\texttt{merge}}, \text{\texttt{conflict}}\}$ specifies the consolidation strategy, $idxs \subseteq \{1, \dots, K_m\}$ are the indices of target entries in $\tilde{\mathcal{S}}_q$, and $k_{\mu}$ is the potential consolidated content. 

The final consolidation operation follows three branching cases: \texttt{new} inserts $(k_q, \mathbf{u}_q)$ as a distinct entry when no overlap is detected; \texttt{merge} supersedes entries indexed by $idxs$ with unified $(k_{\mu}, f_{\text{emb}}(k_{\mu}))$ when the insight complements existing knowledge; \texttt{conflict} purges outdated entries and replaces with $(k_q, \mathbf{u}_q)$ when $k_q$ invalidates previous knowledge.

Algorithm~\ref{alg:nemori-distill} in the Appendix summarizes \textsc{Nemori}'s memory construction pipeline.

\subsection{Response Generation}
\label{subsec:retrieval}
The inference-time use of memory is largely orthogonal to its construction procedures in Section~\ref{subsec: Episodic} and Section~\ref{subsec: Semantic}. Therefore, response generation can in principle accommodate diverse retrieval strategies; here we present the direct setting used in our experiments.

Given a user query $Q$ and its embedding $\mathbf{v}_Q \leftarrow f_{\text{emb}}(Q)$, we retrieve in parallel the top-$k$ episodic entries from the episodic database $\mathcal{D}_e$:
\begin{equation*}
    \tilde{\mathcal{R}}_e = \{(c_i, N_i, P_i, \mathbf{v}_i)\}_{i=1}^{k} \leftarrow \operatorname{Search}(\mathcal{D}_e, \mathbf{v}_Q, k),
\end{equation*}
and the top-$m$ semantic entries from the semantic database $\mathcal{D}_s$:
\begin{equation*}
    \tilde{\mathcal{R}}_s = \{(s_j, \mathbf{u}_j)\}_{j=1}^{m} \leftarrow \operatorname{Search}(\mathcal{D}_s, \mathbf{v}_Q, m).
\end{equation*}
Both $\tilde{\mathcal{R}}_e$ and $\tilde{\mathcal{R}}_s$ are ordered by decreasing similarity to the query embedding $\mathbf{v}_Q$.

The final context for response generation is the concatenation of narrative episodes $\mathcal{R}_e = \{N_i\}_{i=1}^{k}$, raw episodes $\mathcal{R}_p=\{P_d\}_{d=1}^{r}$, and semantic knowledge $\mathcal{R}_s = \{s_j\}_{j=1}^{m}$:
\begin{equation*}
    a \leftarrow f_{\text{LLM}} (\mathcal{P}_{\text{ans}} \parallel Q \parallel \mathcal{R}_e \parallel \mathcal{R}_p \parallel \mathcal{R}_s ),
\end{equation*}
where $\mathcal{P}_{\text{ans}}$ instructs the LLM to generate a response to the question grounded in the retrieved context.
  
Algorithm~\ref{alg:nemori-response} in the Appendix summarizes the response generation procedure described above. 

\begin{table*}[t]
\centering
\resizebox{\textwidth}{!}{%
\begin{tabular}{cl|ccc|ccc|ccc|ccc|ccc}
\toprule
& \multirow{2}{*}{\textbf{Method}} & \multicolumn{3}{c|}{\textbf{Temporal Reasoning}} & \multicolumn{3}{c|}{\textbf{Open Domain}} & \multicolumn{3}{c|}{\textbf{Multi-Hop}} & \multicolumn{3}{c|}{\textbf{Single-Hop}} & \multicolumn{3}{c}{\textbf{Average}} \\
\cmidrule{3-17}
& & $\uparrow$LLM & $\uparrow$F1 & $\uparrow$BLEU & $\uparrow$LLM & $\uparrow$F1 & $\uparrow$BLEU & $\uparrow$LLM & $\uparrow$F1 & $\uparrow$BLEU & $\uparrow$LLM & $\uparrow$F1 & $\uparrow$BLEU & $\uparrow$LLM & $\uparrow$F1 & $\uparrow$BLEU \\
\midrule
\multirow{9}{*}{\rotatebox{90}{\textbf{gpt-4.1-mini}}} 
& Full Context & 74.2 & 47.5 & 40.0 & 56.6 & 28.4 & 22.2 & 77.2 & 44.2 & 33.7 & 86.9 & 61.4 & 53.4 & 80.6 & 53.3 & 45.0 \\
\cmidrule(lr){2-17}
& RAG-4096 & 27.4 & 22.3 & 19.1 & 28.8 & 17.9 & 13.9 & 31.7 & 20.1 & 12.8 & 35.9 & 25.8 & 22.0 & 32.9 & 23.5 & 19.2 \\
& LangMem & 50.8 & \underline{48.5} & \underline{40.9} & 59.0 & \underline{32.8} & \underline{26.4} & \underline{71.0} & \underline{41.5} & \underline{32.5} & \underline{84.5} & \underline{51.0} & \underline{43.6} & \underline{73.4} & \underline{47.6} & \underline{40.0} \\
& Zep & 60.2 & 23.9 & 20.0 & 43.8 & 24.2 & 19.3 & 53.7 & 30.5 & 20.4 & 66.9 & 45.5 & 40.0 & 61.6 & 36.9 & 30.9 \\
& Mem0 & 56.9 & 39.2 & 33.2 & 47.9 & 23.7 & 17.7 & 68.2 & 40.1 & 30.3 & 71.4 & 48.6 & 42.0 & 66.3 & 43.5 & 36.5 \\
& A-MEM & \underline{66.7} & 40.3 & 33.7 & 37.5 & 13.4 & 12.7 & 55.7 & 30.4 & 20.0 & 64.0 & 45.0 & 39.8 & 61.4 & 39.4 & 33.2 \\
& MemoryOS & 37.7 & 36.5 & 27.4 & \underline{60.4} & 30.2 & 25.6 & 62.4 & 34.0 & 25.8 & 68.9 & 44.2 & 37.5 & 60.6 & 39.9 & 32.5 \\
\cmidrule(lr){2-17}
& \textbf{\textsc{Nemori}} & \textbf{77.3} & \textbf{58.7} & \textbf{50.7} & \textbf{56.3} & \textbf{31.7} & \textbf{25.1} & \textbf{74.8} & \textbf{40.8} & \textbf{31.7} & \textbf{87.0} & \textbf{55.7} & \textbf{49.5} & \textbf{80.8} & \textbf{52.1} & \textbf{45.0} \\
& \textit{Improv.} & $\uparrow$15.9\% & $\uparrow$21.0\% & $\uparrow$24.0\% & -- & -- & -- & $\uparrow$5.4\% & -- & -- & $\uparrow$3.0\% & $\uparrow$9.2\% & $\uparrow$13.5\% & $\uparrow$10.1\% & $\uparrow$9.5\% & $\uparrow$12.5\% \\
\midrule
\multirow{9}{*}{\rotatebox{90}{\textbf{gpt-4o-mini}}} 
& Full Context & 56.2 & 44.1 & 36.1 & 48.6 & 24.5 & 17.2 & 66.8 & 35.4 & 26.1 & 83.0 & 53.1 & 44.7 & 72.3 & 46.2 & 37.8 \\
\cmidrule(lr){2-17}
& RAG-4096 & 23.7 & 19.5 & 15.7 & 32.6 & 19.0 & 13.5 & 31.3 & 18.6 & 11.7 & 32.0 & 22.2 & 18.6 & 30.2 & 20.8 & 16.4 \\
& LangMem & 24.9 & 31.9 & 26.2 & \underline{47.6} & \underline{29.4} & \underline{23.5} & 52.4 & 33.5 & 23.9 & 61.4 & 38.8 & 33.1 & 51.3 & 35.8 & 29.4 \\
& Zep & \underline{58.9} & \underline{44.8} & \underline{38.1} & 39.6 & 22.9 & 15.7 & 50.5 & 27.5 & 19.3 & 63.2 & 39.7 & 33.7 & 58.5 & 37.5 & 30.9 \\
& Mem0 & 50.4 & 44.4 & 37.6 & 40.6 & 27.1 & 19.4 & \underline{60.3} & 34.3 & \underline{25.2} & \underline{68.1} & \underline{44.4} & \underline{37.7} & \underline{61.3} & \underline{41.5} & \underline{34.2} \\
& A-MEM & 54.2 & 38.1 & 33.8 & 22.9 & 9.0 & 8.6 & 43.6 & 24.0 & 18.8 & 58.2 & 35.6 & 29.2 & 52.5 & 32.4 & 27.0 \\
& MemoryOS & 38.0 & 38.5 & 27.5 & 45.8 & 26.0 & 19.2 & 52.5 & \underline{35.2} & 24.1 & 62.5 & 43.7 & 37.7 & 54.5 & 39.9 & 31.9 \\
\cmidrule(lr){2-17}
& \textbf{\textsc{Nemori}} & \textbf{67.6} & \textbf{57.3} & \textbf{47.6} & \textbf{45.8} & \textbf{23.9} & \textbf{18.5} & \textbf{61.7} & \textbf{38.1} & \textbf{26.0} & \textbf{81.9} & \textbf{54.8} & \textbf{43.8} & \textbf{73.0} & \textbf{50.3} & \textbf{39.7} \\
& \textit{Improv.} & $\uparrow$14.8\% & $\uparrow$27.9\% & $\uparrow$24.9\% & -- & -- & -- & $\uparrow$2.3\% & $\uparrow$8.2\% & $\uparrow$3.2\% & $\uparrow$20.3\% & $\uparrow$23.4\% & $\uparrow$16.2\% & $\uparrow$19.1\% & $\uparrow$21.2\% & $\uparrow$16.1\% \\
\bottomrule
\end{tabular}%
}
\caption{Performance comparison on LoCoMo dataset. \underline{Underline}: strongest memory system (excluding \textsc{Nemori}). \textit{Improv.}: \textsc{Nemori}'s relative improvement (\%) over strongest memory system.}
\label{tab:main_results}
\end{table*}

\section{Experiments}
We conduct experiments addressing: (RQ1) performance comparison, (RQ2) efficiency analysis, (RQ3) component sensitivity, (RQ4) retrieval configurations, (RQ5) third-party management integration, and (RQ6) scalability to longer contexts.

\subsection{Experimental Setup}
\label{subsec:setup}
\paragraph{Datasets.}
We evaluate \textsc{Nemori} on two distinct benchmarks. 
\textbf{LoCoMo} \citep{maharana-etal-2024-evaluating}: 10 dialogues with 24K average tokens, featuring 1,540 questions across four reasoning categories. 
\textbf{LongMemEval$_\text{S}$} \citep{DBLP:conf/iclr/WuWYZCY25}: 500 conversations with 105K average tokens. While structurally similar to LoCoMo, it presents significantly greater challenges through longer, more realistic conversational contexts, allowing us to assess scalability under demanding conditions.

\paragraph{Baselines.}
We benchmark against seven methods: 
\textbf{Full Context} (entire dialogue history), 
\textbf{RAG-4096} (\citealp{DBLP:conf/nips/LewisPPPKGKLYR020}, 4096-token chunks for dense retrieval), 
\textbf{LangMem} (\citealp{langchain}, automatic knowledge extraction across sessions), 
\textbf{Zep} (\citealp{DBLP:journals/corr/abs-2501-13956}, fact extraction into knowledge graphs), 
\textbf{Mem0} (\citealp{DBLP:journals/corr/abs-2504-19413}, vector-based preference capture), 
\textbf{A-MEM} (\citealp{xu2025amem}, structured notes with evolving links), and 
\textbf{MemoryOS} (\citealp{kang-etal-2025-memory}, hierarchical OS-inspired storage). 

\paragraph{Evaluation Metrics.}
On the LoCoMo dataset, our primary evaluation metric is the \textbf{LLM-judge score} (abbreviated as \textbf{LLM} for simplicity), using gpt-4o-mini as the judge. We additionally report \textbf{F1} and \textbf{BLEU-1}. For the LongMemEval$_\text{S}$ dataset, we also use the LLM-judge score, but with prompts adapted to its task-specific question-answering format, following Zep \citep{DBLP:journals/corr/abs-2501-13956}. These metrics are accuracy metrics of different standard, and are scaled to the 0--100 range, with higher values indicating better performance and 100 denoting a perfect score.

\paragraph{Implementation Details.}
To ensure fair comparison, Mem0 and Zep utilize their commercial APIs to retrieve memory contexts, which are then fed to gpt-4o-mini and gpt-4.1-mini for answer generation. All other methods, including \textsc{Nemori}, employ gpt-4o-mini and gpt-4.1-mini as both internal backbone models and answer generation models. For \textsc{Nemori} specifically, embeddings are generated with text-embedding-3-small. Key hyperparameters were configured as follows: similarity threshold $\tau=0.70$, distillation parameters $K_e=K_m=5, K_s=10$. For retrieval count, we fix $m=2k$; in main experiments $k=10$ (thus $m=20$), while $k$ varies from 2 to 30 in RQ3. To balance informativeness and efficiency, only the top-2 episodic memories include their original conversation text (i.e, $r=2$), as higher-similarity episodes tend to be more informative.

\subsection{Main Results (RQ1)}
Table~\ref{tab:main_results} presents the main performance comparison on LoCoMo. 
Regarding this table, we highlight the following observations:

\paragraph{Strong Performance.}
\textsc{Nemori} achieves the strongest average performance. 
In terms of the average LLM-judge score, with gpt-4.1-mini, \textsc{Nemori} achieves \textbf{80.8}, surpassing LangMem (73.4) by \textbf{10.1\%}; with gpt-4o-mini, it reaches \textbf{73.0}, exceeding Mem0 (61.3) by \textbf{19.1\%}.
Importantly, \textsc{Nemori} slightly exceeds Full Context on both models (80.8 vs.\ 80.6 and 73.0 vs.\ 72.3), suggesting that it effectively captures the intrinsic properties of interaction sequences and recognizes useful experience. 

\paragraph{Exceptional Temporal Reasoning.}
\textsc{Nemori} excels in Temporal Reasoning, achieving LLM-judge score of 77.3 with gpt-4.1-mini (+15.9\% over A-MEM) and 67.6 with gpt-4o-mini (+14.8\% over Zep). 
This result suggests the effectiveness of \textsc{Nemori}'s \emph{episode-centric} design (\S\ref{subsec: Episodic}, \S\ref{subsec: Semantic}), which front-loads part of the reasoning burden from response generation to memory formation and better aligns with the inherent logical structure of experience. 
A case study is provided in Appendix~\ref{app:case_temporal}. 

\textsc{Nemori}'s performance is slightly lower compared to the strongest method on Open Domain. Questions in this category usually require both memory and the backbone model's prior knowledge. See Appendix~\ref{app:open_domain} for further discussion.

\subsection{Efficiency Analysis (RQ2)}
\label{subsec:efficiency}
The efficiency analysis of the memory system can be divided into two stages: memory construction and response generation, where the former corresponds to the distillation and management stages and the latter corresponds to the retrieval stage in the sense of Section~\ref{sec:intro} and Table~\ref{tab:memory_taxonomy}. 

\paragraph{Memory Construction.}
\begin{table}[t]
    \centering
    \resizebox{\columnwidth}{!}{%
    \begin{tabular}{@{}lrrrrr@{}}
    \toprule
    \textbf{Method} & $\uparrow$ \textbf{LLM} & $\downarrow$ \textbf{Calls} & $\downarrow$ \textbf{Input} & $\downarrow$ \textbf{Output} & $\downarrow$ \textbf{Total} \\
    & \textbf{(\%)} & & \textbf{(k)} & \textbf{(k)} & \textbf{(k)} \\
    \midrule
    LangMem & 51.3 & \underline{920.6} & 898.3 & \underline{112.0} & 1010.2 \\
    Mem0 & \underline{61.3} & 1602.2 & 1483.4 & 210.0 & 1693.4 \\
    A-MEM & 52.5 & 1175.5 & 912.6 & 236.8 & 1149.4 \\
    MemoryOS & 54.5 & 1016.1 & \underline{404.5} & 122.0 & \underline{526.5} \\
    \midrule
    \textbf{\textsc{Nemori}} & \textbf{73.0} & \textbf{373.2} & \textbf{277.2} & \textbf{45.7} & \textbf{322.9} \\
    \textit{Improv.} & $\uparrow$19.1\% & $\downarrow$59.5\% & $\downarrow$31.5\% & $\downarrow$59.2\% & $\downarrow$38.7\% \\
    \bottomrule
    \end{tabular}
    }
    \caption{Comparison of memory construction cost on LoCoMo using gpt-4o-mini. 
    The last three columns report token consumption.}
    \label{tab:construction_efficiency}
\end{table}

Table~\ref{tab:construction_efficiency} reports the cost of memory construction on LoCoMo with gpt-4o-mini, where the baseline results are taken from \citet{DBLP:journals/corr/abs-2510-18866} and \textsc{Nemori} is evaluated under the same scope for fair comparison.
\textsc{Nemori} is more cost-efficient than the baselines, reducing LLM calls by \textbf{59.5\%} and token consumption by \textbf{38.7\%}. This result may seem surprising, as \textsc{Nemori} appears to employ a complex pipeline with multiple specialized prompt types. A key reason is that \textsc{Nemori} avoids the trap of \emph{message-wise} processing that many baselines fall into by using \emph{episode} as its primary processing unit.
A finer-grained breakdown of \textsc{Nemori}'s costs is provided in Table~\ref{tab:construction_breakdown} of Appendix.

\paragraph{Response Generation.}
\begin{table}[t]
    \centering
    \resizebox{\columnwidth}{!}{%
    \begin{tabular}{@{}lrrrr@{}}
    \toprule
    \textbf{Method} & \textbf{LLM} & \textbf{Tokens} & \textbf{Search (ms)} & \textbf{Total (ms)} \\
    \midrule
    FullContext & 72.3 & 23,653 & -- & 5,806 \\
    RAG-4096 & 30.2 & 3,430 & 544 & 2,884 \\
    LangMem & 51.3 & 125 & 19,829 & 22,082 \\
    Zep & 58.5 & 2,247 & 522 & 3,255 \\
    Mem0 & 61.3 & 1,027 & 784 & 3,539 \\
    A-MEM & 52.5 & 2,614 & 947 & 2,867 \\
    MemoryOS & 54.5 & 1,560 & 9,910 & 15,220 \\
    \midrule
    \textbf{\textsc{Nemori}} & \textbf{73.0} & 2,745 & 787 & 3,053 \\
    \bottomrule
    \end{tabular}
    }
    \caption{Comparison of response generation cost on LoCoMo using gpt-4o-mini. Search denotes memory retrieval time; Total denotes end-to-end latency from receiving the question to completing the answer.}
    \label{tab:efficiency}
\end{table}
Table~\ref{tab:efficiency} reports retrieval-time efficiency on LoCoMo with gpt-4o-mini. Here, \textsc{Nemori} uses 2,745 tokens on average, an \textbf{88\%} reduction compared with Full Context's 23,653 tokens, while achieving slightly higher accuracy (73.0 vs.\ 72.3) and \textbf{47\%} lower total latency (3,053ms vs.\ 5,806ms). 

\subsection{Ablation Study (RQ3)}
\label{subsec:ablation}
Table~\ref{tab:ablation} presents the ablation results. A finer-grained result is provided in Table~\ref{tab:ablation_full} of Appendix. We highlight the following observations:

\paragraph{Prediction-error-based vs. Direct Distillation.}
The superior performance of \textbf{w/o e} over \textbf{\textsc{Nemori}-s} confirms the effectiveness of \textsc{Nemori}'s adaptive distillation design. 
Both ablations discard the episodic database at response generation. The key difference is that \textbf{w/o e} generates responses using the semantic database from \textsc{Nemori}'s prediction-error-based distillation, while \textbf{\textsc{Nemori}-s} implements direct knowledge distillation over each incoming raw episode with the prompt in Appendix~\ref{app:direct}. 
On gpt-4o-mini, prediction-error-based distillation achieves 65.0 vs. 52.0 for direct distillation (\textbf{+25.0\%}); on gpt-4.1-mini, 74.9 vs. 65.5 (\textbf{+14.4\%}). 

\begin{table}[t]
\centering
\footnotesize
\resizebox{\columnwidth}{!}{%
\begin{tabular}{@{}cll|ccc@{}}
\toprule
& \textbf{Configuration} & \textbf{Mgmt} & \textbf{LLM} & \textbf{F1} & \textbf{BLEU} \\
\midrule
\multirow{8}{*}{\rotatebox{90}{\footnotesize\textbf{gpt-4o-mini}}}
& w/o \textsc{Nemori} & -- & 0.6 & 0.5 & 0.9 \\
& \multirow{2}{*}{Nemori-s} & \ding{51} & 51.7 & 36.4 & 28.9 \\
& & \ding{55} & 52.0 & 36.6 & 29.1 \\
& \multirow{2}{*}{w/o e} & \ding{51} & 64.6 & 46.2 & 37.1 \\
& & \ding{55} & 65.0 & 46.2 & 36.9 \\
& w/o s & \ding{51} & 54.7 & 39.6 & 31.7 \\
& w/o p & \ding{51} & 68.0 & 47.4 & 36.8 \\
& \textbf{\textsc{Nemori}} & \ding{51} & \textbf{73.0} & \textbf{50.3} & \textbf{39.7} \\
\midrule
\multirow{8}{*}{\rotatebox{90}{\footnotesize\textbf{gpt-4.1-mini}}}
& w/o \textsc{Nemori} & -- & 1.2 & 1.6 & 1.5 \\
& \multirow{2}{*}{Nemori-s} & \ding{51} & 66.0 & 41.4 & 34.9 \\
& & \ding{55} & 65.5 & 41.1 & 34.1 \\
& \multirow{2}{*}{w/o e} & \ding{51} & 74.7 & 48.2 & 40.9 \\
& & \ding{55} & 74.9 & 48.1 & 40.7 \\
& w/o s & \ding{51} & 76.9 & 50.0 & 42.9 \\
& w/o p & \ding{51} & 75.7 & 48.1 & 40.9 \\
& \textbf{\textsc{Nemori}} & \ding{51} & \textbf{80.8} & \textbf{52.1} & \textbf{45.0} \\
\bottomrule
\end{tabular}
}
\caption{Ablation study on LoCoMo. w/o \textsc{Nemori} = without \textsc{Nemori}; Nemori-s = semantic-only (direct distillation); w/o e = without episodic retrieval; w/o s = without semantic retrieval; w/o p = without adaptive partitioning (fixed 20-message chunks); \textsc{Nemori} = full framework. Mgmt: \ding{51} = with native management, \ding{55} = naive RAG detailed in Appendix~\ref{app:naiverag}.}
\label{tab:ablation}
\end{table}

\begin{figure}[t]
\centering
\includegraphics[width=\columnwidth]{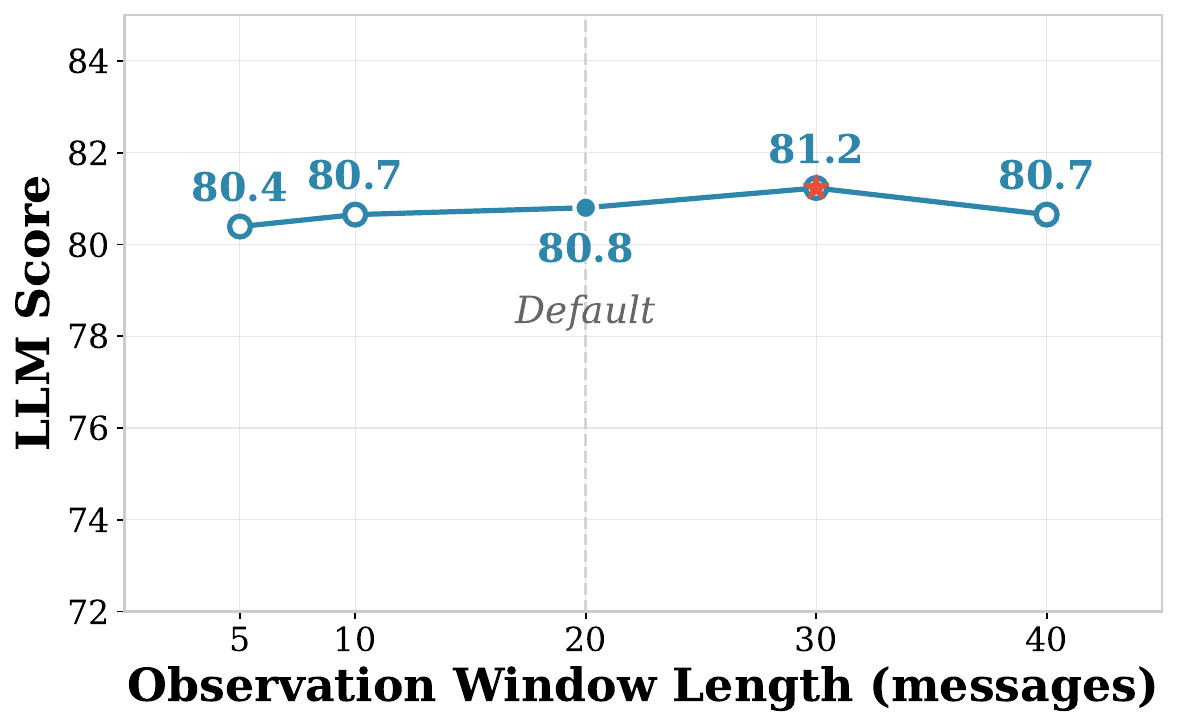}
\caption{Perfomance on LoCoMo with gpt-4.1-mini across observation window lengths $w=5-40$, where $w=20$ is the default setting in the main experiments.}
\label{fig:window_ablation}
\end{figure}

\paragraph{Native Management Contribution.}
Toggling native management (\ding{51} vs.\ \ding{55}) shows minimal impact: 64.6 vs.\ 65.0 on gpt-4o-mini and 74.7 vs.\ 74.9 on gpt-4.1-mini. 
This is expected, as LoCoMo rarely involves knowledge updates requiring consolidation. 
We nevertheless retain native management for real-world deployment where such updates might be common.

\paragraph{Episodic--Semantic Complementarity.}
Both memory types are indispensable.
Removing episodic retrieval (\textbf{w/o e}) drops performance from 73.0 to 65.0 ($-$11.0\%) on gpt-4o-mini and from 80.8 to 74.9 ($-$7.3\%) on gpt-4.1-mini. 
Removing semantic retrieval (\textbf{w/o s}) drops performance from 73.0 to 54.7 ($-$25.1\%) on gpt-4o-mini and from 80.8 to 76.9 ($-$4.8\%) on gpt-4.1-mini.

\paragraph{Observation Window Length.}
As shown in Figure~\ref{fig:window_ablation}, \textsc{Nemori}'s performance remains stable across observation window lengths from 5 to 40, indicating that its design of message partitioning with integration is robust to this hyperparameter.
A finer-grained result is provided in Table~\ref{tab:window_ablation} of Appendix. 

\begin{figure}[t]
\vspace{0.8em}
\centering
\includegraphics[width=\columnwidth]{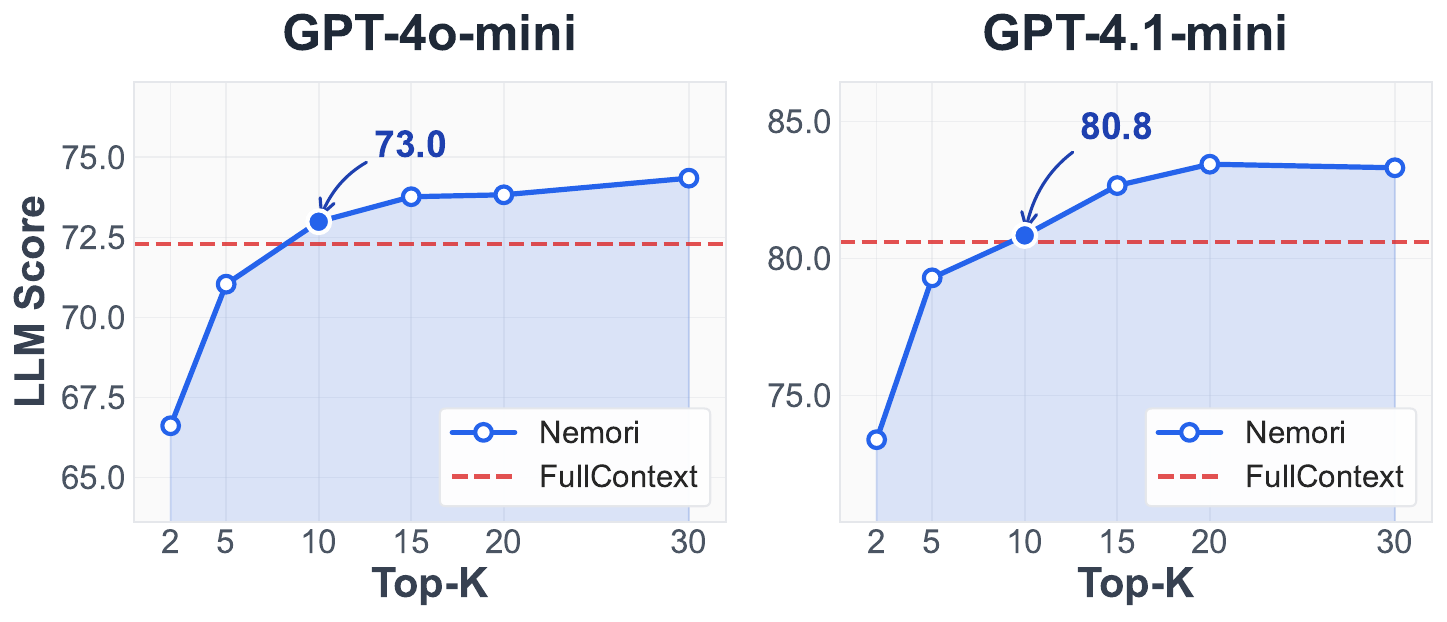}
\caption{Effect of retrieval count $k$ on LLM Score for gpt-4o-mini (left) and gpt-4.1-mini (right). Dashed lines indicate the Full Context baseline. The annotated points mark the default setting ($k{=}10$) used in main experiments.}
\label{fig:topk_ablation}
\vspace{0.2em}
\end{figure}

\begin{table}[t]
\centering
\resizebox{\columnwidth}{!}{%
\begin{tabular}{@{}cc|ccc@{}}
\toprule
\textbf{Index} & \textbf{Retrieve} & \textbf{LLM} & \textbf{F1} & \textbf{BLEU} \\
\midrule
\textbf{N} & \textbf{N} & \textbf{76.9} & \textbf{50.0} & \textbf{42.9} \\
P & N & 76.4 & 50.7 & 43.7 \\
\cmidrule(lr){1-5}
N & P & 77.0 & 50.2 & 42.6 \\
P & P & 75.3 & 50.1 & 42.7 \\
\bottomrule
\end{tabular}
}
\caption{Retrieval strategy ablation on LoCoMo using gpt-4.1-mini. N = narrative episodes; P = raw (partitioned) episodes. \textit{Index} denotes the embedding source; \textit{Retrieve} denotes the content returned to the LLM. When Retrieve = N, the top-2 narratives additionally include their raws, as described in Section~\ref{subsec:setup}. Bold row marks the default setting.}
\label{tab:retrieval_ablation}
\end{table}

\subsection{Retrieval Hyperparameter Analysis (RQ4)}
We analyze two aspects of the retrieval configuration described in Section~\ref{subsec:retrieval}: the sensitivity to retrieval count $k$, and the choice of index--retrieve strategy.

\paragraph{Top-K Sensitivity.}
Figure~\ref{fig:topk_ablation} shows that performance rises sharply as $k$ increases from 2 to 10, then plateaus at a stable level that exceeds \textbf{Full Context}. 
This suggests that a simple Top-K search strategy with mild retrieval count already saturates performance, indicating that \textsc{Nemori} effectively mitigates memory noise through its distillation process. 
A finer-grained result is provided in Table~\ref{tab:topk_ablation} of Appendix. 

\paragraph{Index--Retrieve Strategy.}
Table~\ref{tab:retrieval_ablation} supports the representation prior and our implementation in Section~\ref{sec:narrative}: holding the retrieved content fixed, narrative episode embeddings consistently outperform raw episode embeddings (76.9 vs.\ 76.4 when retrieving N; 77.0 vs.\ 75.3 when retrieving P).
A finer-grained breakdown is provided in Table~\ref{tab:retrieval_ablation_full} in the Appendix.

\begin{table}[t]
\centering
\resizebox{\columnwidth}{!}{%
\begin{tabular}{@{}cl|c|cc|r@{}}
\toprule
& \multirow{2}{*}{\textbf{System}} & \multirow{2}{*}{\textbf{Input}} & \multicolumn{2}{c|}{$\uparrow$\textbf{LLM Score}} & \multirow{2}{*}{$\downarrow$\textbf{MemTokens}} \\
\cmidrule{4-5}
& & & Average & Core & \\
\midrule
\multirow{6}{*}{\rotatebox{90}{\scriptsize\textbf{gpt-4o-mini}}}
& \multirow{3}{*}{A-MEM} 
& P & 52.5 & 52.6 & 397K \\
& & $\mathcal{K}$ & 50.9 & 55.8 & 142K \\
& & $\Delta$ & $\downarrow$3.0\% & $\uparrow$6.1\% & $\downarrow$64.3\% \\
\cmidrule{2-6}
& \multirow{3}{*}{MemoryOS} 
& P & 54.6 & 59.2 & 405K \\
& & $\mathcal{K}$ & 54.0 & 60.3 & 190K \\
& & $\Delta$ & $\downarrow$1.1\% & $\uparrow$1.9\% & $\downarrow$53.1\% \\
\midrule
\multirow{6}{*}{\rotatebox{90}{\scriptsize\textbf{gpt-4.1-mini}}}
& \multirow{3}{*}{A-MEM} 
& P & 61.4 & 60.4 & 498K \\
& & $\mathcal{K}$ & 59.0 & 64.1 & 243K \\
& & $\Delta$ & $\downarrow$3.9\% & $\uparrow$6.1\% & $\downarrow$51.3\% \\
\cmidrule{2-6}
& \multirow{3}{*}{MemoryOS} 
& P & 60.7 & 66.9 & 354K \\
& & $\mathcal{K}$ & 61.4 & 69.2 & 194K \\
& & $\Delta$ & $\uparrow$1.2\% & $\uparrow$3.4\% & $\downarrow$45.3\% \\
\bottomrule
\end{tabular}
}
\caption{Third-party management comparison. P=raw messages; $\mathcal{K}$=\textsc{Nemori}'s distilled semantic knowledge. Core=weighted average excluding Temporal.}
\label{tab:management_comparison}
\end{table}

\subsection{Third-Party Integration (RQ5)}
Table~\ref{tab:management_comparison} evaluates \textsc{Nemori} as an adaptive distillation kernel for third-party memory systems. 
Fed with semantic knowledge $\mathcal{K}$ instead of raw episodes (messages) $P$, both A-MEM and MemoryOS reduce storage by 45--64\% while maintaining average performance ($\pm$4\%), with core scores improving (+1.9\% to +6.1\%). 
A finer-grained breakdown is provided in Table~\ref{tab:input_comparison_perf} and Table~\ref{tab:input_comparison_storage} in the Appendix. 

\subsection{Scalability Analysis (RQ6)}
We evaluate on LongMemEval$_\text{S}$ (105K), an order of magnitude longer than LoCoMo (9K). 
As shown in Table~\ref{tab:longmemeval}, \textsc{Nemori} outperforms Full Context by $+$16.7\% on gpt-4o-mini and $+$13.7\% on gpt-4.1-mini, a substantial increase from the marginal gains on LoCoMo ($+$1.0\% and $+$0.2\%).
The consistent improvements across both datasets suggest that \textsc{Nemori} captures general properties of interaction sequences rather than artifacts of a particular benchmark. 
The widening gap further demonstrates that distillation becomes increasingly valuable as context grows: Full Context suffers from attention dilution over long inputs, while \textsc{Nemori} maintains focused retrieval with 95--96\% fewer tokens. 
\begin{table}[t]
\centering
\resizebox{\columnwidth}{!}{%
\begin{tabular}{@{}cl|cc@{}}
\toprule
& \textbf{Question Type} & \textbf{Full-context} & \textbf{\textsc{Nemori}} \\
& & \textit{(101K tok.)} & \textit{(3.7--4.8K tok.)} \\
\midrule
\multirow{7}{*}{\rotatebox{90}{\footnotesize\textbf{gpt-4o-mini}}}
& Single-session Preference & 6.7 & \textbf{46.7} \\
& Single-session Assistant & \textbf{89.3} & 83.9 \\
& Temporal Reasoning & 42.1 & \textbf{61.7} \\
& Multi-session & 38.3 & \textbf{51.1} \\
& Knowledge Update & \textbf{78.2} & 61.5 \\
& Single-session User & 78.6 & \textbf{88.6} \\
\cmidrule(lr){2-4}
& \textit{Average} & \textit{55.0} & \textit{\textbf{64.2}} \\
\midrule
\multirow{7}{*}{\rotatebox{90}{\footnotesize\textbf{gpt-4.1-mini}}}
& Single-session Preference & 16.7 & \textbf{86.7} \\
& Single-session Assistant & \textbf{98.2} & 92.9 \\
& Temporal Reasoning & 60.2 & \textbf{72.2} \\
& Multi-session & 51.1 & \textbf{55.6} \\
& Knowledge Update & 76.9 & \textbf{79.5} \\
& Single-session User & 85.7 & \textbf{90.0} \\
\cmidrule(lr){2-4}
& \textit{Average} & \textit{65.6} & \textit{\textbf{74.6}} \\
\bottomrule
\end{tabular}
}
\caption{Performance comparison on LongMemEval\textsubscript{S} dataset. \textsc{Nemori} achieves higher accuracy while using 95--96\% less context.}
\label{tab:longmemeval}
\end{table}

\section{Conclusion}
Inspired by cognitive ideas, we propose \textsc{Nemori}, a training-free framework that adaptively assesses the future utility of an agent's experience at the distillation stage, where prediction error deserves retention as memory. 
Guided by three priors about structure, representation and distillation of interaction sequences, \textsc{Nemori}'s cascading modules work in coordination. Its episode-centric design enhances token efficiency, and its management-agnostic design allows it to serve as a distillation layer for downstream memory systems. 
Our results suggest that being agentic need not imply being heuristic: observations over data properties can establish a data-driven space, increasingly important as agents requiring long-term behavioral consistency usually operate beyond human curation.

\section*{Limitations}
We acknowledge two limitations. 
\emph{First}, \textsc{Nemori} focuses on distillation and adopts naive strategies for management and retrieval. This simplicity suffices for the benchmarks studied here, but may become a bottleneck for tasks that demand more sophisticated reasoning over memory. The performance we report should be understood as reflecting this design scope rather than the ceiling of what a complete memory system could achieve.
\emph{Second}, the interfaces we define are currently conceptual, and due to the lack of standardized protocols in the area, concrete integration still requires case-by-case implementation. 

\section*{Acknowledgments}
The authors would first like to thank the anonymous reviewers for their constructive feedback. 
We are also grateful to Shanda Group for providing the resources that supported this work. 
We thank Prof.\ Weiguo Zheng for his advice on highlighting our core contributions, Prof.\ Jiaye Teng for his advice on the paper's overall structure, and Prof.\ Yixuan Qiu for his detailed suggestions on our presentation. 
We also thank Huaqing Zhang and Chenrui Wang for their generous help in reviewing an earlier version of this paper. 

Wenquan Ma thanks the transferable knowledge and skills he gained in Prof.\ Zheng's group and in his previous research under Prof.\ Teng's supervision. 
Jiayan Nan thanks Yize Chen for the support and help in his work and research. 
Finally, the authors thank each other, and look forward to more opportunities for collaboration in the future. 

\bibliography{custom}
\appendix

\section{Implementations of Management}
\label{app: Instantiations}
This appendix details three instantiations of the management interfaces defined in Section~\ref{subsec: Semantic}, demonstrating \textsc{Nemori}'s architectural flexibility.

\subsection{Conceptual Model: Flat Summarization}
This configuration only serves as a conceptual illustration to aid understanding. Here, $\mathcal{M}$ maintains a single monolithic summary rather than a structured database.

\paragraph{Context Evocation.}
The global summary $\mathcal{S}_{sum}$ is returned as constant context:
\begin{equation*}
\operatorname{Evoke}(M_{in}, \mathcal{M}): \quad \mathcal{S}_{in} \leftarrow \mathcal{S}_{sum}
\end{equation*}

\paragraph{Knowledge Consolidation.}
New insights are directly merged into the summary:
\begin{equation*}
\mathcal{S}_{sum} \leftarrow f_{\text{LLM}}(\mathcal{P}_{\text{sum}} \parallel \mathcal{K}_{in} \parallel \mathcal{S}_{sum})
\end{equation*}
where $\mathcal{P}_{\text{sum}}$ instructs the LLM to merge distilled insights into the summary.

\subsection{Variant: Naive RAG}
\label{app:naiverag}
This configuration serves as an ablation in Section~\ref{subsec:ablation}, applying no management to semantic memory. Distilled insights are directly stored and retrieved via similarity search, without conflict resolution.

\paragraph{Context Evocation.}
For each input $M_{in}$, the interface retrieves top-$K_s$ semantically similar entries:
\begin{equation*}
\mathcal{S}_{in} \leftarrow \operatorname{Top-}K_s \left( S_r \in \mathcal{D}_s \mid \operatorname{sim}(\mathbf{v}_{in}, \mathbf{u}_r) > \tau \right)
\end{equation*}

\paragraph{Knowledge Consolidation.}
Each distilled insight $k_q \in \mathcal{K}_{in}$ is simply embedded and appended:
\begin{equation*}
\mathcal{D}_s \leftarrow \mathcal{D}_s \cup \{(k_q, f_{\text{emb}}(k_q))\}
\end{equation*}
No management like conflict detection or merging is performed. This contrasts with our native implementation (Section~\ref{subsec: Semantic}), which applies the full consolidation logic with new/merge/conflict resolution. 

\subsection{External Integration: Third-Party Systems}
\textsc{Nemori} can be implemented with third-party management systems \citep{xu2025amem, kang-etal-2025-memory} by intercepting their context assembly and injecting distilled content.

\paragraph{Context Evocation.}
Most of the memory systems fundamentally operate by conditioning response generation on related context. 
We intercept this context buffer $\tilde{\mathcal{B}}$, assembled by the host's management logic for query $M_{in}$, and repurpose it as the basis for prediction: instead of generating a response, we use it to synthesize an anticipatory schema of what should have occurred:
\begin{equation*}
\mathcal{S}_{in} \leftarrow \tilde{\mathcal{B}}
\end{equation*}

\paragraph{Knowledge Consolidation.}
Each distilled insight $k_q \in \mathcal{K}_{in}$ is injected as an independent message into the host's input sequence, allowing the external system to manage it natively. 
Notably, most of the memory systems discussed in this paper are designed to process explicit factual knowledge, making \textsc{Nemori}'s distilled semantic memory a suitable input. 

\begin{algorithm}[t]
\caption{\textsc{Nemori} Memory Distillation}
\label{alg:nemori-distill}
\begin{algorithmic}[1]
\Require Message buffer $\mathcal{B}_t = \{m_1, \dots, m_z\}$
\Ensure Updated episodic database $\mathcal{D}_e$, semantic database $\mathcal{D}_s$
\Statex \textit{\% --- Episodic Memory Integration (\S\ref{subsec: Episodic}) ---}
\State Partition $\mathcal{B}_t$ into raw episodes $\mathbf{P} = \{P_1, \dots, \allowbreak P_n\} \leftarrow f_{\text{LLM}}(\allowbreak\mathcal{P}_{\text{par}} \parallel \mathcal{B}_t)$
\For{each raw episode $P_j \in \mathbf{P}$}
    \State Generate narrative and cue $(N_j, c_j) \leftarrow f_{\text{LLM}}(\mathcal{P}_{\text{nar}} \parallel P_j)$
    \State Compute embedding $\mathbf{v}_j \leftarrow f_{\text{emb}}(c_j \parallel N_j)$
    \State Retrieve candidates from $\mathcal{D}_e$ and decide merge-or-insert
    \State Obtain episodic memory $M_{in}$ (merged $M_{\nu}$ or new $M_j$)
    \Statex \textit{\% --- Semantic Knowledge Distillation (\S\ref{subsec: Semantic}) ---}
    \State Evoke context $\mathcal{S}_{in} \leftarrow \operatorname{Evoke}(M_{in}, \mathcal{M})$
    \State Synthesize anticipatory schema $\hat{P}_{in} \leftarrow f_{\text{LLM}}(\mathcal{P}_{\text{ant}} \parallel c_{in} \parallel \mathcal{S}_{in})$
    \State Distill semantic insights $\mathcal{K}_{in} \leftarrow f_{\text{LLM}}(\allowbreak \mathcal{P}_{\text{dis}} \allowbreak \parallel P_{in} \parallel \hat{P}_{in})$
    \State $\operatorname{Consolidate}(\mathcal{K}_{in}, \mathcal{M})$
\EndFor
\end{algorithmic}
\end{algorithm}

\begin{algorithm}[t]
\caption{\textsc{Nemori} Response Generation}
\label{alg:nemori-response}
\begin{algorithmic}[1]
\Require Query $Q$, episodic database $\mathcal{D}_e$, semantic database $\mathcal{D}_s$
\Ensure Response $a$
\State Compute query embedding $\mathbf{v}_Q \leftarrow f_{\text{emb}}(Q)$
\State Retrieve $\tilde{\mathcal{R}}_e \leftarrow \operatorname{Search}(\mathcal{D}_e, \mathbf{v}_Q, k)$; extract $\mathcal{R}_e = \{N_i\}_{i=1}^{k}$, $\mathcal{R}_p = \{P_d\}_{d=1}^{r}$
\State Retrieve $\tilde{\mathcal{R}}_s \leftarrow \operatorname{Search}(\mathcal{D}_s, \mathbf{v}_Q, m)$; extract $\mathcal{R}_s = \{s_j\}_{j=1}^{m}$
\State Generate response $a \leftarrow f_{\text{LLM}}(\mathcal{P}_{\text{ans}} \parallel Q \parallel \mathcal{R}_e \parallel \mathcal{R}_p \parallel \mathcal{R}_s)$
\State \Return $a$
\end{algorithmic}
\end{algorithm}

\section{Case Study}
This section presents two representative cases from the main text, highlighting how \textsc{Nemori} supports temporal reasoning and open-domain question answering.

\subsection{Temporal Reasoning}
\label{app:case_temporal}
To illustrate how \textsc{Nemori} enhances response quality, we provide a representative case from the LoCoMo dataset. 

\textbf{Question:} \textit{``When did Jon receive mentorship?''}

\textbf{Challenge:} The original conversation contains relative temporal references like ``yesterday'' without explicit dates, requiring temporal reasoning.

\textbf{Full Context baseline:} Confused by the term ``yesterday'' in the raw dialogue, the model incorrectly answered with the conversation date (June 16).

\textbf{\textsc{Nemori}:} Retrieved both the relevant episodic memory (preserving conversational context) and a semantic memory that had already distilled the temporal information into explicit fact: \textit{``Jon was mentored on June 15, 2023.''} By combining episodic context with pre-reasoned semantic knowledge, \textsc{Nemori} transforms complex reasoning into simple fact retrieval.

\textbf{Insight:} This demonstrates the capability of ``reasoning during memory formation.'' The prediction error highlight that the specific date is unexpected given prior knowledge, prompting its distillation as semantic memory. 

\subsection{Open Domain}
\label{app:open_domain}
On the Open Domain subset, \textsc{Nemori}'s LLM score is slightly below the strongest memory system baseline, with gaps of 6.8\% under gpt-4.1-mini (56.3 vs. 60.4) and 3.8\% under gpt-4o-mini (45.8 vs. 47.6). We note that this subset is not a pure measure of the memory procedure's effectiveness, specifically:

In LoCoMo, many such questions are not directly answerable from the original conversation history alone; instead, they require the backbone model to recognize a conversational description and map it to an item of general world knowledge. As a result, performance in this category depends not only on memory quality, but also on the model's prior knowledge.

A representative example is the question: \textit{``What is the game with different colored cards that John was talking about with James?''} The gold answer is \textit{``UNO''}, but the dialogue itself never explicitly names UNO. Instead, the transcript only states that the players discussed a game with multi-colored cards and matching by color or number, while also noting that the speaker had forgotten its name. Accordingly, \textsc{Nemori}'s episodic memory preserves this conversational evidence, and the semantic memory distills the same game description, but neither memory can inject the missing lexical label if it is absent from the interaction history. In such cases, whether the final answer becomes \textit{``UNO''} depends largely on the backbone model's ability to recognize the description from prior knowledge, rather than on a failure of memory distillation or retrieval.

\section{Additional Experiment Results}
\label{app:additional_results}
This appendix provides detailed experimental results that supplement the main paper. All experiments use the setup described in Section~\ref{subsec:setup}. 
\begin{table}[t]
    \centering
    \resizebox{\columnwidth}{!}{%
    \begin{tabular}{@{}lrrrr@{}}
    \toprule
    \textbf{Component} & \textbf{Input (k)} & \textbf{Output (k)} & \textbf{Total (k)} & \textbf{Ratio} \\
    \midrule
    Partition (\S \ref{sec:partitioning}) & 44.7 & 4.6 & 49.3 & 15.3\% \\
    Narration (\S \ref{sec:narrative}) & 99.8 & 23.9 & 123.6 & 38.3\% \\
    Integration (\S \ref{sec:integration}) & 43.8 & 8.4 & 52.2 & 16.2\% \\
    Distillation (\S \ref{subsec: Semantic}) & 88.9 & 8.8 & 97.7 & 30.3\% \\
    \bottomrule
    \end{tabular}%
    }
    \caption{Component-wise breakdown of \textsc{Nemori}'s memory construction cost on LoCoMo with gpt-4o-mini.}
    \label{tab:construction_breakdown}
\end{table}

\begin{table*}[p]
\centering
\resizebox{\textwidth}{!}{%
\begin{tabular}{cll|ccc|ccc|ccc|ccc|ccc}
\toprule
& \multirow{2}{*}{\textbf{Configuration}} & \multirow{2}{*}{\textbf{Mgmt}} & \multicolumn{3}{c|}{\textbf{Temporal Reasoning}} & \multicolumn{3}{c|}{\textbf{Open Domain}} & \multicolumn{3}{c|}{\textbf{Multi-Hop}} & \multicolumn{3}{c|}{\textbf{Single-Hop}} & \multicolumn{3}{c}{\textbf{Overall}} \\
\cmidrule{4-18}
& & & LLM & F1 & BLEU & LLM & F1 & BLEU & LLM & F1 & BLEU & LLM & F1 & BLEU & LLM & F1 & BLEU \\
\midrule
\multirow{7}{*}{\rotatebox{90}{\footnotesize\textbf{gpt-4o-mini}}}
& \multirow{2}{*}{Nemori-s} & \ding{51} & 33.3 & 36.8 & 31.1 & 49.0 & 24.4 & 18.6 & 47.9 & 30.5 & 20.2 & 60.3 & 39.7 & 32.1 & 51.7 & 36.4 & 28.9 \\
& & \ding{55} & 32.7 & 35.9 & 30.4 & 40.6 & 21.8 & 17.0 & 47.5 & 31.3 & 20.7 & 62.1 & 40.3 & 32.7 & 52.0 & 36.6 & 29.1 \\
& \multirow{2}{*}{w/o e} & \ding{51} & 57.9 & 53.0 & 44.6 & 53.1 & 26.5 & 19.6 & 57.8 & 35.5 & 24.4 & 70.8 & 49.4 & 40.4 & 64.6 & 46.2 & 37.1 \\
& & \ding{55} & 56.7 & 52.8 & 44.8 & 54.2 & 27.8 & 20.6 & 59.9 & 36.5 & 24.8 & 71.1 & 48.9 & 39.8 & 65.0 & 46.2 & 36.9 \\
& w/o s & \ding{51} & 32.7 & 38.9 & 33.0 & 42.7 & 22.2 & 17.0 & 53.9 & 33.0 & 22.2 & 64.7 & 44.1 & 36.1 & 54.7 & 39.6 & 31.7 \\
& w/o p & \ding{51} & 56.7 & 52.7 & 43.4 & 45.8 & 25.0 & 19.3 & 59.9 & 36.3 & 23.5 & 77.5 & 51.7 & 40.7 & 68.0 & 47.4 & 36.8 \\
& \textbf{\textsc{Nemori}} & \ding{51} & \textbf{67.6} & \textbf{57.3} & \textbf{47.6} & \textbf{45.8} & \textbf{23.9} & \textbf{18.5} & \textbf{61.7} & \textbf{38.1} & \textbf{26.0} & \textbf{81.9} & \textbf{54.8} & \textbf{43.8} & \textbf{73.0} & \textbf{50.3} & \textbf{39.7} \\
\midrule
\multirow{7}{*}{\rotatebox{90}{\footnotesize\textbf{gpt-4.1-mini}}}
& \multirow{2}{*}{Nemori-s} & \ding{51} & 46.4 & 42.2 & 33.7 & 49.0 & 26.5 & 20.7 & 67.4 & 36.2 & 28.8 & 74.9 & 44.5 & 39.0 & 66.0 & 41.4 & 34.9 \\
& & \ding{55} & 47.0 & 42.5 & 32.5 & 50.0 & 28.5 & 22.7 & 70.6 & 38.9 & 29.9 & 72.7 & 42.8 & 37.4 & 65.5 & 41.1 & 34.1 \\
& \multirow{2}{*}{w/o e} & \ding{51} & 63.2 & 49.6 & 41.1 & 52.1 & 27.2 & 21.2 & 72.7 & 38.9 & 29.1 & 82.4 & 53.2 & 47.0 & 74.7 & 48.2 & 40.9 \\
& & \ding{55} & 65.4 & 51.0 & 42.5 & 56.3 & 29.1 & 23.0 & 70.2 & 39.0 & 29.5 & 82.2 & 52.2 & 45.8 & 74.9 & 48.1 & 40.7 \\
& w/o s & \ding{51} & 73.5 & 54.3 & 46.9 & 55.2 & 26.9 & 21.3 & 73.1 & 41.9 & 32.5 & 81.9 & 53.7 & 47.3 & 76.9 & 50.0 & 42.9 \\
& w/o p & \ding{51} & 67.3 & 53.4 & 45.0 & 52.1 & 24.7 & 19.5 & 71.3 & 40.1 & 30.7 & 83.1 & 51.4 & 45.2 & 75.7 & 48.1 & 40.9 \\
& \textbf{\textsc{Nemori}} & \ding{51} & \textbf{77.3} & \textbf{58.7} & \textbf{50.7} & \textbf{56.3} & \textbf{31.7} & \textbf{25.1} & \textbf{74.8} & \textbf{40.8} & \textbf{31.7} & \textbf{87.0} & \textbf{55.7} & \textbf{49.5} & \textbf{80.8} & \textbf{52.1} & \textbf{45.0} \\
\bottomrule
\end{tabular}%
}
\caption{Category-wise Ablation study on LoCoMo. Nemori-s = semantic-only (direct distillation); w/o e = without episodic retrieval; w/o s = without semantic retrieval; w/o p = without adaptive partitioning (fixed 20-message chunks); \textsc{Nemori} = full framework. Mgmt: \ding{51} = with native management, \ding{55} = naive RAG detailed in Section~\ref{app:naiverag}.}
\label{tab:ablation_full}
\end{table*}
\begin{table*}[p]
\centering
\resizebox{\textwidth}{!}{%
\begin{tabular}{c|ccc|ccc|ccc|ccc|ccc}
\toprule
\multirow{2}{*}{$w$} & \multicolumn{3}{c|}{\textbf{Temporal}} & \multicolumn{3}{c|}{\textbf{Open Domain}} & \multicolumn{3}{c|}{\textbf{Multi-Hop}} & \multicolumn{3}{c|}{\textbf{Single-Hop}} & \multicolumn{3}{c}{\textbf{Overall}} \\
\cmidrule{2-16}
& LLM & F1 & BLEU & LLM & F1 & BLEU & LLM & F1 & BLEU & LLM & F1 & BLEU & LLM & F1 & BLEU \\
\midrule
5 & 77.0 & 57.6 & 49.6 & 59.4 & 30.0 & 24.9 & 73.8 & 43.0 & 34.0 & 86.3 & 55.0 & 48.5 & 80.4 & 51.8 & 44.6 \\
10 & 77.6 & 59.0 & 50.4 & 57.3 & 29.5 & 24.7 & 77.0 & 44.0 & 34.3 & 85.7 & 54.5 & 48.1 & 80.7 & 52.0 & 44.6 \\
\textbf{20} & \textbf{77.3} & \textbf{58.7} & \textbf{50.7} & \textbf{56.3} & \textbf{31.7} & \textbf{25.1} & \textbf{74.8} & \textbf{40.8} & \textbf{31.7} & \textbf{87.0} & \textbf{55.7} & \textbf{49.5} & \textbf{80.8} & \textbf{52.1} & \textbf{45.0} \\
30 & 76.6 & 57.7 & 49.6 & 60.4 & 32.2 & 26.1 & 79.4 & 45.0 & 34.8 & 86.0 & 55.4 & 48.8 & 81.2 & 52.5 & 45.0 \\
40 & 76.3 & 58.6 & 50.4 & 54.2 & 27.1 & 21.4 & 77.3 & 42.8 & 34.0 & 86.4 & 55.0 & 48.4 & 80.7 & 51.8 & 44.5 \\
\bottomrule
\end{tabular}%
}
\caption{Performance across different observation window lengths on LoCoMo dataset with gpt-4.1-mini. w=20 (bold) is the default setting used in main experiments.}
\label{tab:window_ablation}
\end{table*}

\begin{table*}[p]
\centering
\resizebox{\textwidth}{!}{%
\begin{tabular}{@{}cl|ccc|ccc|ccc|ccc|ccc@{}}
\toprule
& \multirow{2}{*}{$k$} & \multicolumn{3}{c|}{\textbf{Temporal}} & \multicolumn{3}{c|}{\textbf{Open Domain}} & \multicolumn{3}{c|}{\textbf{Multi-Hop}} & \multicolumn{3}{c|}{\textbf{Single-Hop}} & \multicolumn{3}{c}{\textbf{Overall}} \\
\cmidrule{3-17}
& & LLM & F1 & BLEU & LLM & F1 & BLEU & LLM & F1 & BLEU & LLM & F1 & BLEU & LLM & F1 & BLEU \\
\midrule
\multirow{6}{*}{\rotatebox{90}{\footnotesize\textbf{gpt-4o-mini}}}
& 2 & 62.3 & 55.3 & 46.5 & 41.7 & 20.9 & 15.4 & 55.0 & 33.7 & 21.8 & 75.0 & 51.1 & 40.9 & 66.6 & 46.9 & 37.0 \\
& 5 & 64.5 & 56.7 & 47.4 & 47.9 & 24.9 & 19.2 & 62.4 & 36.7 & 25.0 & 79.1 & 53.3 & 42.4 & 71.0 & 49.2 & 38.8 \\
& \textbf{10} & \textbf{67.6} & \textbf{57.3} & \textbf{47.6} & \textbf{45.8} & \textbf{23.9} & \textbf{18.5} & \textbf{61.7} & \textbf{38.1} & \textbf{26.0} & \textbf{81.9} & \textbf{54.8} & \textbf{43.8} & \textbf{73.0} & \textbf{50.3} & \textbf{39.7} \\
& 15 & 68.9 & 58.6 & 48.4 & 44.8 & 24.5 & 19.0 & 61.4 & 37.0 & 25.4 & 83.1 & 54.9 & 43.6 & 73.8 & 50.5 & 39.8 \\
& 20 & 67.9 & 57.4 & 47.7 & 45.8 & 24.3 & 18.9 & 63.5 & 38.2 & 25.9 & 82.8 & 54.7 & 43.1 & 73.8 & 50.4 & 39.4 \\
& 30 & 68.2 & 58.9 & 48.4 & 45.8 & 23.9 & 18.8 & 64.5 & 37.7 & 25.6 & 83.2 & 55.0 & 43.3 & 74.4 & 50.7 & 39.6 \\
\midrule
\multirow{6}{*}{\rotatebox{90}{\footnotesize\textbf{gpt-4.1-mini}}}
& 2 & 68.5 & 52.8 & 45.5 & 52.1 & 25.7 & 20.5 & 64.5 & 38.2 & 28.3 & 80.6 & 51.7 & 45.6 & 73.4 & 47.8 & 40.9 \\
& 5 & 74.1 & 56.7 & 48.9 & 55.2 & 28.8 & 22.6 & 73.1 & 41.9 & 32.3 & 86.1 & 54.4 & 48.1 & 79.3 & 51.0 & 43.8 \\
& \textbf{10} & \textbf{77.3} & \textbf{58.7} & \textbf{50.7} & \textbf{56.3} & \textbf{31.7} & \textbf{25.1} & \textbf{74.8} & \textbf{40.8} & \textbf{31.7} & \textbf{87.0} & \textbf{55.7} & \textbf{49.5} & \textbf{80.8} & \textbf{52.1} & \textbf{45.0} \\
& 15 & 80.1 & 59.6 & 51.7 & 57.3 & 31.1 & 24.3 & 77.7 & 42.6 & 33.2 & 88.2 & 55.9 & 49.5 & 82.7 & 52.7 & 45.4 \\
& 20 & 79.8 & 59.4 & 51.1 & 58.3 & 30.3 & 24.4 & 81.2 & 45.3 & 35.5 & 88.5 & 55.6 & 49.1 & 83.4 & 52.9 & 45.5 \\
& 30 & 80.4 & 60.0 & 52.0 & 60.4 & 29.9 & 23.0 & 79.4 & 44.3 & 35.0 & 88.4 & 56.1 & 49.5 & 83.3 & 53.1 & 45.7 \\
\bottomrule
\end{tabular}
}
\caption{Category-wise breakdown of retrieval count $k$ on LoCoMo for gpt-4o-mini and gpt-4.1-mini. Semantic memory count is fixed at $m = 2k$. Bold rows mark the default setting ($k{=}10$) used in main experiments.}
\label{tab:topk_ablation}
\end{table*}
\begin{table*}[p]
\centering
\resizebox{\textwidth}{!}{%
\begin{tabular}{cc|ccc|ccc|ccc|ccc|ccc}
\toprule
\multirow{2}{*}{\textbf{Index}} & \multirow{2}{*}{\textbf{Retrieve}} & \multicolumn{3}{c|}{\textbf{Temporal}} & \multicolumn{3}{c|}{\textbf{Open Domain}} & \multicolumn{3}{c|}{\textbf{Multi-Hop}} & \multicolumn{3}{c|}{\textbf{Single-Hop}} & \multicolumn{3}{c}{\textbf{Overall}} \\
\cmidrule{3-17}
& & LLM & F1 & BLEU & LLM & F1 & BLEU & LLM & F1 & BLEU & LLM & F1 & BLEU & LLM & F1 & BLEU \\
\midrule
\textbf{N} & \textbf{N} & \textbf{72.9} & \textbf{55.7} & \textbf{48.3} & \textbf{55.2} & \textbf{26.0} & \textbf{20.4} & \textbf{72.3} & \textbf{41.8} & \textbf{32.8} & \textbf{82.3} & \textbf{53.3} & \textbf{47.1} & \textbf{76.9} & \textbf{50.0} & \textbf{42.9} \\
P & N & 71.7 & 56.5 & 49.0 & 47.9 & 24.0 & 19.6 & 72.3 & 40.5 & 30.8 & 82.8 & 55.0 & 48.7 & 76.4 & 50.7 & 43.7 \\
\cmidrule(lr){1-17}
N & P & 73.8 & 43.4 & 35.8 & 55.2 & 27.3 & 21.5 & 72.0 & 43.2 & 32.4 & 82.4 & 57.8 & 51.0 & 77.0 & 50.2 & 42.6 \\
P & P & 69.5 & 42.5 & 35.2 & 51.0 & 22.6 & 18.2 & 67.0 & 40.2 & 30.5 & 83.0 & 59.4 & 52.4 & 75.3 & 50.1 & 42.7 \\
\bottomrule
\end{tabular}%
}
\caption{Category-wise breakdown of retrieval strategy ablation on LoCoMo using gpt-4.1-mini. N = narrative episodes; P = raw (partitioned) episodes. \textit{Index} denotes the embedding source; \textit{Retrieve} denotes the content returned to the LLM. When Retrieve = N, the top-2 narratives additionally include their raws, as described in Section~\ref{subsec:setup}. Bold row marks the default setting.}
\label{tab:retrieval_ablation_full}
\end{table*}
\begin{table*}[p]
\centering
\resizebox{\textwidth}{!}{%
\begin{tabular}{@{}ll|c|cccc|cc@{}}
\toprule
\textbf{Model} & \textbf{System} & \textbf{Input} & \textbf{Temp} & \textbf{Open} & \textbf{Multi} & \textbf{Single} & \textbf{Average} & \textbf{Core} \\
\midrule
\multirow{6}{*}{gpt-4o-mini}
& \multirow{3}{*}{A-MEM} 
& N & 54.2 & 22.9 & 43.6 & 58.2 & 52.5 & 52.6 \\
& & $\mathcal{K}$ & 33.6 & 38.5 & 50.4 & 59.1 & 50.9 & 55.8 \\
& & $\Delta$ & $\downarrow$38.0\% & $\uparrow$68.1\% & $\uparrow$15.4\% & $\uparrow$1.6\% & $\downarrow$3.0\% & $\uparrow$6.1\% \\
\cmidrule{2-9}
& \multirow{3}{*}{MemoryOS} 
& N & 38.0 & 45.8 & 52.5 & 62.5 & 54.6 & 59.2 \\
& & $\mathcal{K}$ & 30.8 & 44.8 & 58.5 & 62.4 & 54.0 & 60.3 \\
& & $\Delta$ & $\downarrow$18.9\% & $\downarrow$2.2\% & $\uparrow$11.5\% & $\downarrow$0.2\% & $\downarrow$1.1\% & $\uparrow$1.9\% \\
\midrule
\multirow{6}{*}{gpt-4.1-mini}
& \multirow{3}{*}{A-MEM} 
& N & 66.7 & 37.5 & 55.7 & 64.0 & 61.4 & 60.4 \\
& & $\mathcal{K}$ & 41.1 & 41.7 & 58.2 & 68.0 & 59.0 & 64.1 \\
& & $\Delta$ & $\downarrow$38.4\% & $\uparrow$11.1\% & $\uparrow$4.5\% & $\uparrow$6.3\% & $\downarrow$3.9\% & $\uparrow$6.1\% \\
\cmidrule{2-9}
& \multirow{3}{*}{MemoryOS} 
& N & 37.7 & 60.4 & 62.4 & 68.9 & 60.7 & 66.9 \\
& & $\mathcal{K}$ & 32.7 & 58.3 & 62.4 & 72.3 & 61.4 & 69.2 \\
& & $\Delta$ & $\downarrow$13.3\% & $\downarrow$3.5\% & -- & $\uparrow$5.0\% & $\uparrow$1.2\% & $\uparrow$3.4\% \\
\bottomrule
\end{tabular}
}
\caption{LLM Score comparison of third-party management systems with different input sources on LoCoMo. N=raw conversation; $\mathcal{K}$=\textsc{Nemori}'s distilled semantic memory. Core=weighted average excluding Temporal.}
\label{tab:input_comparison_perf}
\end{table*}

\begin{table*}[p]
\centering
\resizebox{\textwidth}{!}{%
\begin{tabular}{@{}ll|c|rrr@{}}
\toprule
\textbf{Model} & \textbf{System} & \textbf{Input} & \textbf{Tokens} & \textbf{Chars} & \textbf{Entries} \\
\midrule
\multirow{6}{*}{gpt-4o-mini}
& \multirow{3}{*}{A-MEM} 
& N & 396,812 & 2,475,511 & 5,882 \\
& & $\mathcal{K}$ & 141,682 & 820,025 & 2,725 \\
& & $\Delta$ & $\downarrow$64.3\% & $\downarrow$66.9\% & $\downarrow$53.7\% \\
\cmidrule{2-6}
& \multirow{3}{*}{MemoryOS} 
& N & 404,611 & 1,956,432 & 3,014 \\
& & $\mathcal{K}$ & 189,662 & 927,678 & 2,613 \\
& & $\Delta$ & $\downarrow$53.1\% & $\downarrow$52.6\% & $\downarrow$13.3\% \\
\midrule
\multirow{6}{*}{gpt-4.1-mini}
& \multirow{3}{*}{A-MEM} 
& N & 498,234 & 3,811,770 & 5,882 \\
& & $\mathcal{K}$ & 242,801 & 1,459,777 & 2,676 \\
& & $\Delta$ & $\downarrow$51.3\% & $\downarrow$61.7\% & $\downarrow$54.5\% \\
\cmidrule{2-6}
& \multirow{3}{*}{MemoryOS} 
& N & 354,463 & 1,712,305 & 3,017 \\
& & $\mathcal{K}$ & 193,744 & 1,023,709 & 2,383 \\
& & $\Delta$ & $\downarrow$45.3\% & $\downarrow$40.2\% & $\downarrow$21.0\% \\
\bottomrule
\end{tabular}
}
\caption{Memory storage comparison of third-party management systems with different input sources. N = raw conversation; $\mathcal{K}$ = \textsc{Nemori}'s distilled semantic memory. \textit{Entries} denotes the number of memory entries defined by each system's own storage format (comparable within but not across systems). \textit{Tokens} and \textit{Chars} are measured by concatenating all entries. $\downarrow$ indicates reduction.}
\label{tab:input_comparison_storage}
\end{table*}

\paragraph{Memory Construction Cost (Table~\ref{tab:construction_breakdown})}
Finer-grained results of Section~\ref{subsec:efficiency}. The main cost comes from the narrative episode generation (38.3\%) and semantic knowledge distillation (30.3\%). 

\paragraph{Ablation Study (Table~\ref{tab:ablation_full}).}
Finer-grained results of Section~\ref{subsec:ablation}. 
Prediction-error-based distillation consistently outperforms direct knowledge distillation across categories. 
The improvement is most pronounced in \textbf{Temporal Reasoning}, from 33.3 to 57.9 (\textbf{+73.9\%}) on gpt-4o-mini and from 46.4 to 63.2 (\textbf{+36.2\%}) on gpt-4.1-mini, where prediction-error-based distillation effectively identifies and transforms time-sensitive information.

\paragraph{Observation Window Length (Table~\ref{tab:window_ablation}).}
Finer-grained results of Figure~\ref{fig:window_ablation}. 
Overall scores remain stable ($\pm$1\%) across window lengths from 5 to 40. 
Category-level variation is likewise small, confirming that \textsc{Nemori}'s design of message partitioning with integration is robust to this hyperparameter.

\paragraph{Top-K Sensitivity (Table~\ref{tab:topk_ablation}).}
Performance rises sharply as $k$ increases from 2 to 10, then plateaus. Strongest average performance is achieved at $k$=15$\sim$20, but $k$=10 provides 97\% of peak performance with lower computational cost.

\paragraph{Retrieval Strategy (Table~\ref{tab:retrieval_ablation_full}).}
The \textbf{N$\rightarrow$P} configuration achieves a marginally higher LLM score (77.0 vs.\ 76.9), as raw text preserves factual details for answer generation. 
We default to \textbf{N$\rightarrow$N} for simplicity, since the difference is negligible and narrative retrieval avoids returning lengthy raw episodes.

\paragraph{Third-Party Management (Tables~\ref{tab:input_comparison_perf} and \ref{tab:input_comparison_storage}).}
Using \textsc{Nemori}'s semantic memory as input reduces storage by \textbf{45--64\%} while improving Core scores by 1.9--6.1\%, demonstrating that distilled memory provides a compact yet information-rich representation suitable for downstream management systems.

\section{Prompt Templates}
\label{app:prompts}
This appendix provides the complete prompt templates used in \textsc{Nemori}'s pipeline. 
\subsection{Core Distillation Prompts}
This subsection presents the prompts for the main distillation modules described in Section~\ref{subsec: Episodic} and Section~\ref{subsec: Semantic}, instantiated with our native management implementation.
\subsubsection{Local Message Partitioning Prompt \texorpdfstring{($\mathcal{P}_{\text{par}}$)}{(P\_par)}}

\begin{promptbox}{Local Message Partitioning Prompt}
   You are an intelligent conversation segmentation expert. Your task is to analyze a batch of messages and group them into coherent episodes.
    
    \vspace{0.5em}
    
    \noindent \textbf{You will receive \{count\} messages numbered from 1 to \{count\}:} \{messages\} \\
    
    \vspace{0.5em}
    
    \noindent \#\# Your Task\\
Analyze these messages and group them into coherent episodes with **HIGH SENSITIVITY** to topic shifts. Be strict and create NEW episodes when detecting:\\

1. **Topic Change** (Highest Priority):\\
   - Do the new messages introduce a completely different topic?\\
   - Is there a shift from one specific event to another?\\
   - Has the conversation moved from one question to an unrelated new question?\\

2. **Intent Transition**:\\
   - Has the purpose of the conversation changed? (e.g., from casual chat to seeking help, from discussing work to discussing personal life)\\
   - Has the core question or issue of the current topic been answered or fully discussed?\\

3. **Temporal Markers**:\\
   - Are there temporal transition markers ("earlier", "before", "by the way", "oh right", "also", etc.)?\\
   - Is the time gap between messages more than 30 minutes?\\

4. **Structural Signals**:\\
   - Are there explicit topic transition phrases ("changing topics", "speaking of which", "quick question", etc.)?\\
   - Are there concluding statements indicating the current topic is finished?\\

5. **Content Relevance**:\\
   - How related is the new message to the previous discussion? (Consider splitting if relevance < 30\%)\\
   - Does it involve completely different people, places, or events?\\

Decision Principles:\\
- **Prioritize topic independence**: Each episode should revolve around one core topic or event\\
- **When in doubt, split**: When uncertain, lean towards starting a new episode\\
- **Maintain reasonable length**: A single episode typically shouldn't exceed 10-15 messages\\

\#\# Output Format\\
Return a JSON object with episodes, where each episode contains:\\
- `indices`: List of message numbers (1-based) belonging to this episode\\
- `topic`: Brief, specific description of what this episode is about\\

Example output:\\
\{\{\\
    "episodes": [\\
        \{\{\\
            "indices": [1, 2, 3, 4],\\
            "topic": "Discussion about weekend hiking plans"\\
        \}\},\\
        \{\{\\
            "indices": [5, 6, 7],\\
            "topic": "Questions about Python programming"\\
        \}\},\\
        \{\{\\
            "indices": [8, 9],\\
            "topic": "Work schedule discussion"\\
        \}\}\\
    ]\\
\}\}\\

\#\# Important Guidelines\\
- Episodes can have non-consecutive indices if messages are interleaved\\
- An episode should typically contain 2-15 messages\\
- Focus on topical coherence over strict chronological order\\
- When in doubt, prefer smaller, more focused episodes\\

Return only the JSON object, no additional text.\\
\end{promptbox}

\subsubsection{Narrative Episode Generation Prompt \texorpdfstring{($\mathcal{P}_{\text{nar}}$)}{(P\_nar)}}

\begin{promptbox}{Narrative Episode Generation Prompt}
    You are an episodic memory generation expert. Please convert the following conversation into an episodic memory.
    
    \vspace{0.5em}
    
    \noindent \textbf{Conversation content:} \{conversation\} \\
    \textbf{Boundary detection reason:} \{boundary\_reason\} \\
    
    \vspace{0.5em}
    
    \noindent Please analyze the conversation to extract time information and generate a structured episodic memory. Return only a JSON object containing the following three fields:\\
\{\{\\
    "episodic\_cue": "A concise, descriptive title that accurately summarizes the theme (10-20 words)",\\
    "narrative\_episode": "A detailed description of the conversation in third-person narrative. It must include all important information: who participated in the conversation at what time, what was discussed, what decisions were made, what emotions were expressed, and what plans or outcomes were formed. Write it as a coherent story so that the reader can clearly understand what happened. Ensure that time information is precise to the hour, including year, month, day, and hour.",\\
    "timestamp": "YYYY-MM-DDTHH:MM:SS format timestamp representing when this episode occurred (analyze from message timestamps or content)"\\
\}\}

Time Analysis Instructions:\\
1. **Primary Source**: Look for explicit timestamps in the message metadata or content\\
2. **Secondary Source**: Analyze temporal references in the conversation content ("yesterday", "last week", "this morning", etc.)\\
3. **Fallback**: If no time information is available, use a reasonable estimate based on context\\
4. **Format**: Always return timestamp in ISO format: "2024-01-15T14:30:00"\\

Requirements:\\
1. The title should be specific and easy to search (including key topics/activities).\\
2. The content must include all important information from the conversation.\\
3. Convert the dialogue format into a narrative description.\\
4. Maintain chronological order and causal relationships.\\
5. Use third-person unless explicitly first-person.\\
6. Include specific details that aid keyword search.\\
7. Notice the time information, and write the time information in the content.\\
8. When relative times (e.g., last week, next month, etc.) are mentioned in the conversation, you need to convert them to absolute dates (year, month, day). Write the converted time in parentheses after the original time reference.\\
9. **IMPORTANT**: Analyze the actual time when the conversation happened from the message timestamps or content, not the current time.\\

Example:\\
If the conversation is about someone planning to go hiking and the messages have timestamps from March 14, 2024 at 3:00 PM:
\{\{\\
    "title": "Weekend Hiking Plan March 16, 2024: Sunrise Trip to Mount Rainier",\\
    "content": "On March 14, 2024 at 3:00 PM, the user expressed interest in going hiking on the upcoming weekend (March 16, 2024) and sought advice. They particularly wanted to see the sunrise at Mount Rainier, having heard the scenery is beautiful. When asked about gear, they received suggestions including hiking boots, warm clothing (as it's cold at the summit), a flashlight, water, and high-energy food. The user decided to leave at 4:00 AM on Saturday, March 16, 2024 to catch the sunrise and planned to invite friends for the adventure. They were very excited about the trip, hoping to connect with nature.",\\
    "timestamp": "2024-03-14T15:00:00"\\
\}\}

Return only the JSON object, do not add any other text:
\end{promptbox}

\subsubsection{Optimal Candidate Identification Prompt \texorpdfstring{($\mathcal{P}_{\text{sel}}$)}{(P\_sel)}}
\begin{promptbox}{Optimal Candidate Identification Prompt}
    You are an episodic memory merge decision expert. Determine if a new episode should be merged with an existing similar episode.
    
    \vspace{0.5em}
    \#\# New Episode\\
    \noindent \textbf{Time Range:} \{new\_time\_range\} \\
    \textbf{Content:} \{new\_content\} \\
    \textbf{Candidate Episodes to Merge With:} \{candidates\}
    
    \vspace{0.5em}
    
    \noindent \#\# Your Task\\
Decide whether the new episode should:\\
1. **merge**: Merge with one of the candidates (they describe the same event/topic)\\
2. **new**: Keep as a separate new episode (it's a distinct event)\\

\#\# Merge Criteria\\
Merge ONLY if:\\
- Both episodes describe the SAME event or conversation session\\
- They have significant temporal overlap or are very close in time\\
- The content is clearly a continuation or different perspective of the same topic\\
- Merging would create a more complete picture without mixing different events\\

Do NOT merge if:\\
- They are different events/conversations even if on similar topics\\
- They are separated by significant time gaps (>1 hour)\\
- They involve different contexts or participants\\

\#\# Output Format\\
Return JSON:\\
\{\{\\
    "decision": "merge" or "new",\\
    "merge\_target\_id": "episode\_id\_to\_merge\_with" (only if decision is "merge", otherwise null),\\
    "reason": "Brief explanation of your decision"\\
\}\}\\

Return only the JSON object, no additional text.\\
\end{promptbox}

\subsubsection{Episodic Integration Prompt \texorpdfstring{($\mathcal{P}_{\text{int}}$)}{(P\_int)}}
\begin{promptbox}{Episodic Integration Prompt}
    You are an episodic memory merge content generator. Combine two related episodes into a single, coherent episode.
    
    \vspace{0.5em}
    \#\#Original Episode
    \noindent \textbf{Time Range:} \{original\_time\_range\} \\
    \textbf{Title:} \{original\_title\} \\
    \textbf{Content:} \{original\_content\}
    
    \#\# New Episode to Merge
    \noindent \textbf{Time Range:} \{new\_time\_range\} \\
    \textbf{Title:} \{new\_title\} \\
    \textbf{Content:} \{new\_content\}
    
    \noindent \textbf{Combined Event Details:} \{combined\_events\} \\
    
    \vspace{0.5em}
    
    \noindent Your Task\\
Generate a merged episode that:\\
1. Combines information from both episodes without duplication\\
2. Maintains chronological flow of events\\
3. Preserves all important details from both episodes\\
4. Creates a coherent narrative\\

\#\# Output Format\\
Return JSON with the merged episode content:\\
\{\{\\
    "title": "Merged episode title that captures the complete topic",\\
    "content": "Detailed narrative combining both episodes chronologically. Include all participants, key decisions, emotions, and outcomes. Use third-person narrative style.",\\
    "timestamp": "ISO format timestamp of when the merged episode occurred (use earliest time)"\\
\}\}\\

\#\# Guidelines\\
- Integrate details naturally, don't just concatenate\\
- Eliminate redundancy while preserving unique information\\
- Maintain temporal coherence in the narrative\\
- Use specific details that aid searchability\\
- Write in third-person narrative style\\

Return only the JSON object, no additional text.\\
\end{promptbox}

\subsubsection{Anticipatory Schema Synthesis Prompt \texorpdfstring{($\mathcal{P}_{\text{ant}}$)}{(P\_ant)}}
\begin{promptbox}{Anticipatory Schema Synthesis Prompt}
    You are a knowledge-based episode prediction system. Your task is to reconstruct a complete conversation episode based on limited clues and your knowledge base.\\

IMPORTANT: You are predicting the ACTUAL CONTENT and KNOWLEDGE of what happened, not the writing style or format.\\

\#\# Input Information

    \vspace{0.5em}
    
    \noindent \textbf{Episodic Cue (Title/Summary):} \{episode\_title\} \\
    \textbf{Evoked Context (Prior Knowledge):} \{evoked\_context\}
    
    \vspace{0.5em}
    
    \noindent \#\# Your Task

Based on the above clues, reconstruct what you believe happened in this episode. Focus on:\\
1. **Core Facts**: What specific information was discussed?\\
2. **Key Decisions**: What choices or conclusions were made?\\
3. **Knowledge Exchange**: What knowledge was shared or learned?\\
4. **Logical Flow**: How did the conversation progress?\\

\#\# What to IGNORE\\
- Writing style or level of detail\\
- Specific formatting or structure\\
- Exact phrasing or word choices\\
- Whether timestamps are included in the text\\
- How formal or casual the language is\\

\#\# Output Format\\

Generate a natural narrative that captures what you predict happened. Write it as if you're describing the episode to someone else. Focus on the SUBSTANCE, not the STYLE.\\

Your prediction:\\
\end{promptbox}

\subsubsection{Prediction Error Distillation Prompt \texorpdfstring{($\mathcal{P}_{\text{dis}}$)}{(P\_dis)}}
\begin{promptbox}{Prediction Error Distillation Prompt}
    You are extracting valuable knowledge by comparing original conversation with predicted content.
    
    \vspace{0.5em}
    
    \noindent \textbf{Actual Episode ($P_{in}$ - Ground Truth):} \{original\_messages\} \\
    \textbf{Anticipatory Schema ($\hat{P}_{in}$ - Expectation):} \{predicted\_episode\} \\
    
    \vspace{0.5em}
    
    \noindent \#\# Your Task:\\
Extract ONLY the valuable knowledge that exists in the original but is missing or misrepresented in the prediction.\\

\#\# What to Extract:\\
Knowledge that is:\\
- Factual and will remain true over time\\
- Specific (names, titles, preferences, reasons)\\
- Useful for future interactions\\
- Not captured accurately in the prediction\\

\#\# What to Ignore:\\
- Temporary states or emotions\\
- Conversational flow or style\\
- Information already well-represented in prediction\\
- Social pleasantries or reactions\\

\#\# Examples:\\

Original: "I'm Alice, a senior engineer at Google. I switched from Java to Python last year because I wanted to work on ML projects."\\
Predicted: "Alice discussed their programming experience."\\
Extract: \\
- "Alice is a senior engineer at Google"\\
- "Alice switched from Java to Python for ML projects"\\

Original: "My favorite book is 'Deep Learning' by Goodfellow. I read it three times because the math explanations are so clear."\\
Predicted: "Alice mentioned liking technical books."\\
Extract:\\
- "Alice's favorite book is 'Deep Learning' by Goodfellow"\\
- "Alice values clear mathematical explanations in technical books"\\

Original: "I've been with Microsoft since 2019, started as a junior developer and got promoted to team lead in 2022. Planning to finish my online CS masters by December 2024."\\
Predicted: "Alice works at Microsoft and is studying."\\
Extract:\\
- "Alice has been at Microsoft since 2019 (5+ years)"\\
- "Alice was promoted from junior developer to team lead in 2022"\\
- "Alice is pursuing an online CS masters degree, expected completion December 2024"\\

\#\# Output Format:\\
\{\{\\
    "statements": [\\
        "First factual statement extracted from the gap",\\
        "Second factual statement extracted from the gap",\\
        "..."\\
    ]\\
\}\}

Important: \\
- Each statement should be self-contained and understandable without context\\
- Use present tense for persistent facts\\
- Include specific names, titles, and details\\
- Focus on quality over quantity - only extract truly valuable knowledge
\end{promptbox}

\subsubsection{Semantic Consolidation Prompt \texorpdfstring{($\mathcal{P}_{\text{con}}$)}{(P\_con)}}
\begin{promptbox}{Semantic Consolidation Prompt}
    You are a conservative knowledge base maintainer. Your default action is NEW unless you are ABSOLUTELY CERTAIN about merging or conflict.
    
    \vspace{0.5em}
    \#\# New Item
    \noindent \textbf{Type:} \{new\_type\} \\
    \textbf{Content:} \{new\_content\} \\
    \textbf{Existing Similar Items:} \{candidates\}
    
    \vspace{0.5em}
    
    \noindent \#\# Actions (choose exactly one)\\
1. **NEW** (DEFAULT): Add the new item. Choose this if:\\
   - The items describe different facts, events, or entities\\
   - The items refer to different times, places, or contexts\\
   - You have ANY doubt about whether they are truly identical or contradictory\\

2. **MERGE** (RARE): Only if the new item and existing item(s) express the EXACT SAME fact with just different wording. Example: "User likes coffee" and "The user enjoys coffee" are merge-able.\\

3. **CONFLICT\_DELETE** (VERY RARE): Only if the new item DIRECTLY CONTRADICTS existing item(s) about the SAME specific fact. Example: "User lives in Beijing" vs "User lives in Shanghai" (same attribute, different value).\\

\#\# Output (valid JSON)\\
- NEW: \{\{"decision": "NEW", "reason": "..."\}\}\\
- MERGE: \{\{"decision": "MERGE", "target\_ids": ["id1"], "new\_content": "canonical phrasing (<=100 words)", "reason": "..."\}\}\\
- CONFLICT\_DELETE: \{\{"decision": "CONFLICT\_DELETE", "target\_ids": ["id1"], "reason": "..."\}\}\\

\#\# CRITICAL RULES\\
- **Default to NEW** - when in doubt, always choose NEW\\
- Similar topics $\neq$ same fact. "User has a cat" and "User has a dog" are BOTH valid, choose NEW\\
- Only MERGE when items are semantically IDENTICAL (just rephrased)\\
- Only CONFLICT\_DELETE for direct contradictions about the SAME attribute\\
- Preserve information richness - losing unique details is worse than having duplicates\\
\end{promptbox}

\subsection{Direct Distillation Prompt (\textsc{Nemori}-s)}
\label{app:direct}
This prompt corresponds to the \textbf{\textsc{Nemori}-s} configuration in our ablation study in Section~\ref{subsec:ablation}, which performs direct knowledge distillation without prediction-error-based distillation.
\begin{promptbox}{Direct Distillation Prompt}
    You are an AI memory system. Extract HIGH-VALUE, PERSISTENT semantic memories from the following episodes.

CRITICAL: Focus on extracting LONG-TERM VALUABLE KNOWLEDGE, not temporary conversation details.

    \vspace{0.5em}
    
    \noindent \textbf{Episodes to analyze:} \{episodes\} 
    
    \vspace{0.5em}
    
    \noindent \#\# HIGH-VALUE Knowledge Criteria

Extract ONLY knowledge that passes these tests:\\
- **Persistence Test**: Will this still be true in 6 months?\\
- **Specificity Test**: Does it contain concrete, searchable information?\\
- **Utility Test**: Can this help predict future user needs?\\
- **Independence Test**: Can be understood without conversation context?\\

\#\# HIGH-VALUE Categories (FOCUS ON THESE):

1. **Identity \& Professional**\\
   - Names, titles, companies, roles\\
   - Education, qualifications, skills\\
   
2. **Persistent Preferences**  \\
   - Favorite books, movies, music, tools\\
   - Technology preferences with reasons\\
   - Long-term likes and dislikes\\
   
3. **Technical Knowledge**\\
   - Technologies used (with versions)\\
   - Architectures, methodologies\\
   - Technical decisions and rationales\\
   
4. **Relationships**\\
   - Names of family, colleagues, friends\\
   - Team structure, reporting lines\\
   - Professional networks\\
   
5. **Goals \& Plans**\\
   - Career objectives\\
   - Learning goals\\
   - Project plans\\
   
6. **Patterns \& Habits**\\
   - Regular activities\\
   - Workflows, schedules\\
   - Recurring challenges\\

\#\# Examples:\\

HIGH-VALUE (Extract these):\\
- "Caroline's favorite book is 'Becoming Nicole' by Amy Ellis Nutt"\\
- "The user works at ByteDance as a senior ML engineer"\\
- "The user prefers PyTorch over TensorFlow for debugging"\\
- "The user's team lead is named Sarah"\\
- "The user is learning Rust for systems programming"\\
- "The user has been practicing yoga since March 2021"\\
- "The user joined Amazon in August 2020 as a data scientist"\\
- "The user plans to relocate to Seattle in January 2025"\\

LOW-VALUE (Skip these):\\
- "The user thanked the assistant"\\
- "The user was confused about X"\\
- "The user appreciated the help"\\
- "The conversation was productive"\\
- Any temporary emotions or reactions\\

\#\# Output Format\\

Return ONLY high-value knowledge in JSON format:\\
\{\{\\
    "statements": [\\
        "First high-value persistent fact...",\\
        "Second high-value persistent fact...",\\
        "Third high-value persistent fact..."\\
    ]\\
\}\}\\

Quality over quantity - extract only knowledge that truly helps understand the user long-term.
\end{promptbox}

\subsection{Response Generation Prompt \texorpdfstring{($\mathcal{P}_{\text{ans}}$)}{(P\_ans)}}
This unified prompt is used for response generation across all evaluation tasks on both LoCoMo and LongMemEval$_\text{S}$ datasets.
\begin{promptbox}{Answer Generation Prompt}
    You are an intelligent memory assistant tasked with retrieving accurate information from conversation memories.\\
    \# CONTEXT:\\
    You have access to memories from two speakers in a conversation. These memories contain\\
    timestamped information that may be relevant to answering the question.\\

    \# INSTRUCTIONS:\\
    1. Carefully analyze all provided memories from both speakers\\
    2. Pay special attention to the timestamps to determine the answer\\
    3. If the question asks about a specific event or fact, look for direct evidence in the memories\\
    4. If the memories contain contradictory information, prioritize the most recent memory\\
    5. If there is a question about time references (like "last year", "two months ago", etc.),
       calculate the actual date based on the memory timestamp. For example, if a memory from
       4 May 2022 mentions "went to India last year," then the trip occurred in 2021.\\
    6. Always convert relative time references to specific dates, months, or years. For example,
       convert "last year" to "2022" or "two months ago" to "March 2023" based on the memory
       timestamp. Ignore the reference while answering the question.\\
    7. Focus only on the content of the memories from both speakers. Do not confuse character
       names mentioned in memories with the actual users who created those memories.\\
    8. The answer should be less than 5-6 words.\\

    \# APPROACH (Think step by step):\\
    1. First, examine all memories that contain information related to the question\\
    2. Examine the timestamps and content of these memories carefully\\
    3. Look for explicit mentions of dates, times, locations, or events that answer the question\\
    4. If the answer requires calculation (e.g., converting relative time references), show your work\\
    5. Formulate a precise, concise answer based solely on the evidence in the memories\\
    6. Double-check that your answer directly addresses the question asked\\
    7. Ensure your final answer is specific and avoids vague time references\\

    \vspace{0.5em}
    
    \noindent \textbf{Episodic Memories:} \{episodic\} \\
    \textbf{Semantic Memories:} \{semantic\} \\
    \textbf{Question:} \{question\}
    
    \vspace{0.5em}
    
    \noindent Answer:
\end{promptbox}

\subsection{LLM-as-Judge Prompts}
\subsubsection{LoCoMo}
LoCoMo uses a single unified evaluation prompt for all question categories.
\begin{promptbox}{LLM-as-Judge Prompt}
    Your task is to label an answer to a question as 'CORRECT' or 'WRONG'. You will be given the following data:\\
    (1) a question (posed by one user to another user),\\ 
    (2) a 'gold' (ground truth) answer, \\
    (3) a generated answer\\
which you will score as CORRECT/WRONG.\\

The point of the question is to ask about something one user should know about the other user based on their prior conversations.\\
The gold answer will usually be a concise and short answer that includes the referenced topic, for example:\\
Question: Do you remember what I got the last time I went to Hawaii?\\
Gold answer: A shell necklace\\
The generated answer might be much longer, but you should be generous with your grading - as long as it touches on the same topic as the gold answer, it should be counted as CORRECT.\\ 

For time related questions, the gold answer will be a specific date, month, year, etc. The generated answer might be much longer or use relative time references (like "last Tuesday" or "next month"), but you should be generous with your grading - as long as it refers to the same date or time period as the gold answer, it should be counted as CORRECT. Even if the format differs (e.g., "May 7th" vs "7 May"), consider it CORRECT if it's the same date.\\

Now it's time for the real question:\\

\vspace{0.5em}
    
    \noindent \textbf{Question:} \{question\} \\
    \textbf{Gold answer:} \{gold\_answer\} \\
    \textbf{Generated answer:} \{generated\_answer\}
    
    \vspace{0.5em}

\noindent First, provide a short (one sentence) explanation of your reasoning, then finish with CORRECT or WRONG. \\
Do NOT include both CORRECT and WRONG in your response, or it will break the evaluation script.\\

Just return the label CORRECT or WRONG in a json format with the key as "label".
\end{promptbox}

\subsubsection{LongMemEval\texorpdfstring{$_\text{S}$}{\_S}}
In contrast to LoCoMo's unified prompt, LongMemEval$_\text{S}$ uses task-specific prompts for evaluation. We present the four variants below.

\begin{promptbox}{Temporal Reasoning Prompt}
I will give you a question, a correct answer, and a response from a model. Please answer yes if the response contains the correct answer. Otherwise, answer no. If the response is equivalent to the correct answer or contains all the intermediate steps to get the correct answer, you should also answer yes. If the response only contains a subset of the information required by the answer, answer no. In addition, do not penalize off-by-one errors for the number of days. If the question asks for the number of days/weeks/months, etc., and the model makes off-by-one errors (e.g., predicting 19 days when the answer is 18), the model's response is still correct.

\vspace{0.5em}

\noindent \textbf{Question:} \{question\} \\
\textbf{Correct Answer:} \{gold\_answer\} \\
\textbf{Response:} \{response\}
\end{promptbox}

\begin{promptbox}{Knowledge Update Prompt}
I will give you a question, a correct answer, and a response from a model. Please answer yes if the response contains the correct answer. Otherwise, answer no. If the response contains some previous information along with an updated answer, the response should be considered as correct as long as the updated answer is the required answer.

\vspace{0.5em}

\noindent \textbf{Question:} \{question\} \\
\textbf{Correct Answer:} \{gold\_answer\} \\
\textbf{Response:} \{response\}
\end{promptbox}

\begin{promptbox}{Single Session Preference Prompt}
I will give you a question, a rubric for desired personalized response, and a response from a model. Please answer yes if the response satisfies the desired response. Otherwise, answer no. The model does not need to reflect all the points in the rubric. The response is correct as long as it recalls and utilizes the user's personal information correctly.

\vspace{0.5em}

\noindent \textbf{Question:} \{question\} \\
\textbf{Rubric:} \{gold\_answer\} \\
\textbf{Response:} \{response\}
\end{promptbox}

\begin{promptbox}{Default Prompt}
I will give you a question, a correct answer, and a response from a model. Please answer yes if the response contains the correct answer. Otherwise, answer no. If the response is equivalent to the correct answer or contains all the intermediate steps to get the correct answer, you should also answer yes. If the response only contains a subset of the information required by the answer, answer no.

\vspace{0.5em}

\noindent \textbf{Question:} \{question\} \\
\textbf{Correct Answer:} \{gold\_answer\} \\
\textbf{Response:} \{response\}
\end{promptbox}

\end{document}